\documentclass[Afour,sageh,times]{sagej}
\def\volumeyear{2019}

\usepackage{tikz} 
\usepackage{tkz-graph}
\usepackage{tkz-berge}
	\usetikzlibrary{backgrounds,fit,shapes,snakes,arrows,shapes.geometric,positioning}
	\usetikzlibrary{intersections,patterns,shapes.misc}
	\usetikzlibrary{decorations.pathmorphing}
	\tikzstyle{block} = [rectangle, rounded corners, minimum width=3cm, minimum height=1cm,text centered, draw=black, fill=red!30]
	\tikzstyle{new} = [rectangle, rounded corners, minimum width=1cm, minimum
	height=1cm,text centered, draw=black, fill=blue!10!white, dashed]
	\tikzstyle{arrow} = [thick,->,>=stealth]
	\usetikzlibrary{calc, quotes}

\usepackage{fontawesome}
\DeclareFontFamily{OT1}{pzc}{}
\DeclareFontShape{OT1}{pzc}{m}{it}{<-> s * [1.200] pzcmi7t}{}
\DeclareMathAlphabet{\mathpzc}{OT1}{pzc}{m}{it}

\AtBeginDocument{%
  \DeclareMathAlphabet\PazoBB{U}{fplmbb}{m}{n}%
}

\usepackage{dsfont}
\usepackage{microtype}
\usepackage{diagbox}
\usepackage{multirow}
\usepackage{dashbox}

\usepackage{sidecap}

\usepackage[framemethod=tikz]{mdframed}
\usepackage{moreverb,url}
\RequirePackage{amsmath}
\usepackage{url}
\usepackage{caption}
\usepackage[justification=centering]{subcaption}
\usepackage{color}
\usepackage{float}
\usepackage{graphicx}

\usepackage{secdot}
\sectiondot{subsection}
\sectiondot{subsubsection}

\newcommand\BibTeX{{\rmfamily B\kern-.05em \textsc{i\kern-.025em b}\kern-.08em
T\kern-.1667em\lower.7ex\hbox{E}\kern-.125emX}}
\newcommand{\norm}[1]{\left\lVert#1\right\rVert}
\newcommand{\Gcal}{\mathcal{G}}
\newcommand{\pitilde}{\tilde{\pi}}
\newcommand{\Pitilde}{\tilde{\Pi}}

\newcommand{\Yedit}[2]{{ #2}}
\graphicspath{{figures/}{revision_figures}}

\begin{document}

\runninghead{Tian et al.}

\title{Search and Rescue under the Forest Canopy using Multiple UAVs}

\author{Yulun Tian\affilnum{1}, Katherine Liu\affilnum{1}, Kyel Ok\affilnum{1},
  Loc Tran\affilnum{2},  Danette Allen\affilnum{2}, 
  Nicholas Roy\affilnum{1} and Jonathan P.\ How\affilnum{1}}

\affiliation{\affilnum{1}Massachusetts Institute of Technology, Cambridge, MA, USA\\
\affilnum{2}NASA Langley Research Center,
Hampton, Virginia, USA}

\corrauth{Yulun Tian}

\email{yulun@mit.edu}

\begin{abstract}

We present a multi-robot system for GPS-denied search and rescue under the forest canopy.
Forests are particularly challenging environments for collaborative exploration and mapping,
in large part due to the existence of severe perceptual aliasing which hinders reliable loop closure detection for mutual localization and map fusion. Our proposed system features unmanned aerial vehicles (UAVs) that perform onboard sensing, estimation, and planning.
When communication is available, each UAV transmits compressed {tree-based} submaps to a central ground station for collaborative simultaneous localization and mapping (CSLAM). To overcome high measurement noise and perceptual aliasing, we use the local configuration of a group of trees as a distinctive feature for robust loop closure detection. Furthermore, we propose a novel procedure based on cycle consistent multiway matching to recover from incorrect pairwise data associations. The returned global data association is guaranteed to be cycle consistent, and is shown to 
improve both precision and recall compared to the input pairwise associations. The proposed multi-UAV system is validated both in simulation and during real-world collaborative exploration missions at NASA Langley Research Center.
\end{abstract}

\maketitle

\section{Introduction}

\begin{figure}[h]
	\centering
    \begin{subfigure}{0.48\textwidth}
	\includegraphics[width=1.0\textwidth]{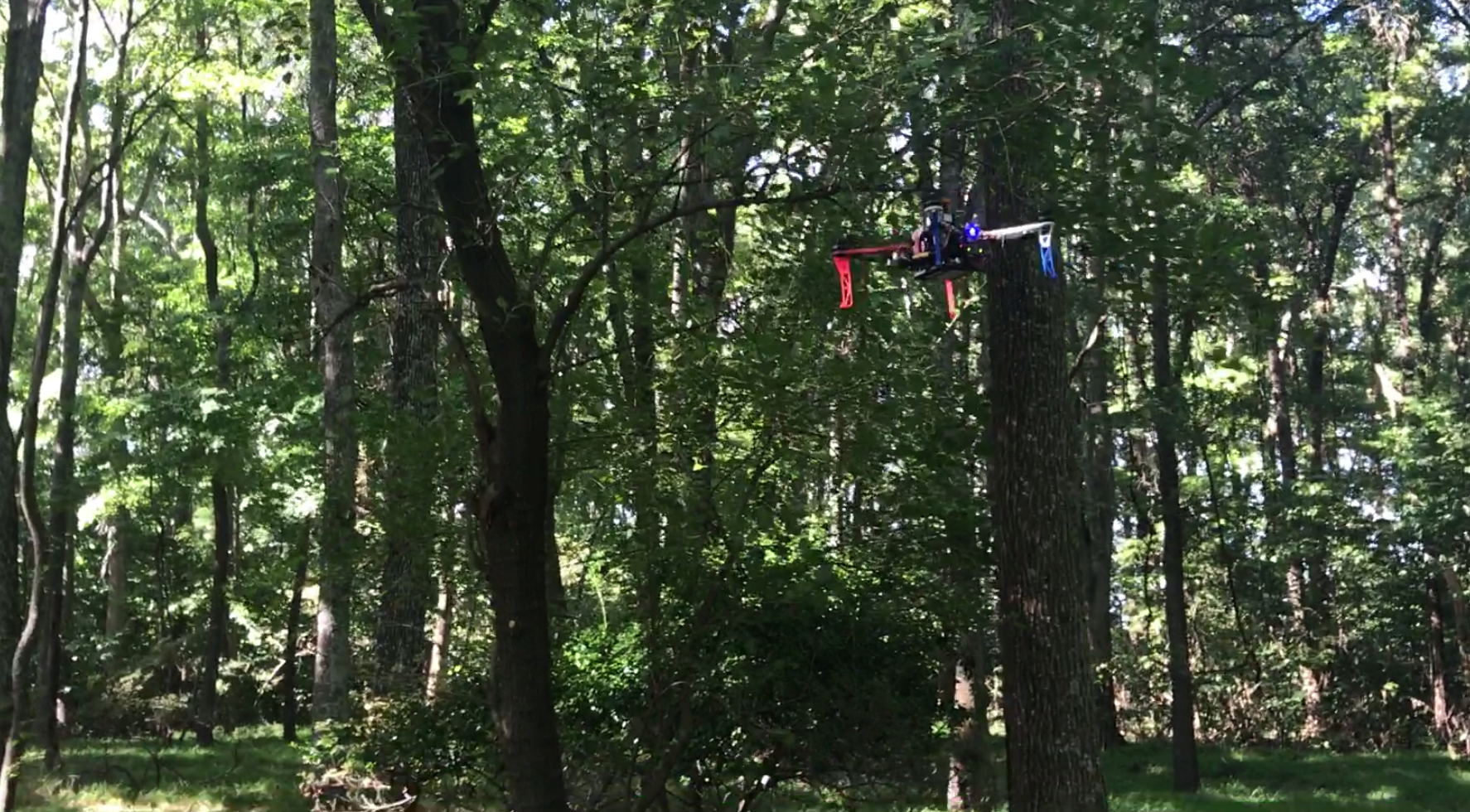}
	\caption{Image of one of our UAVs during a collaborative mapping mission in a forest.}
	\end{subfigure}
	\begin{subfigure}{0.48\textwidth}
	\includegraphics[width=1.0\textwidth]{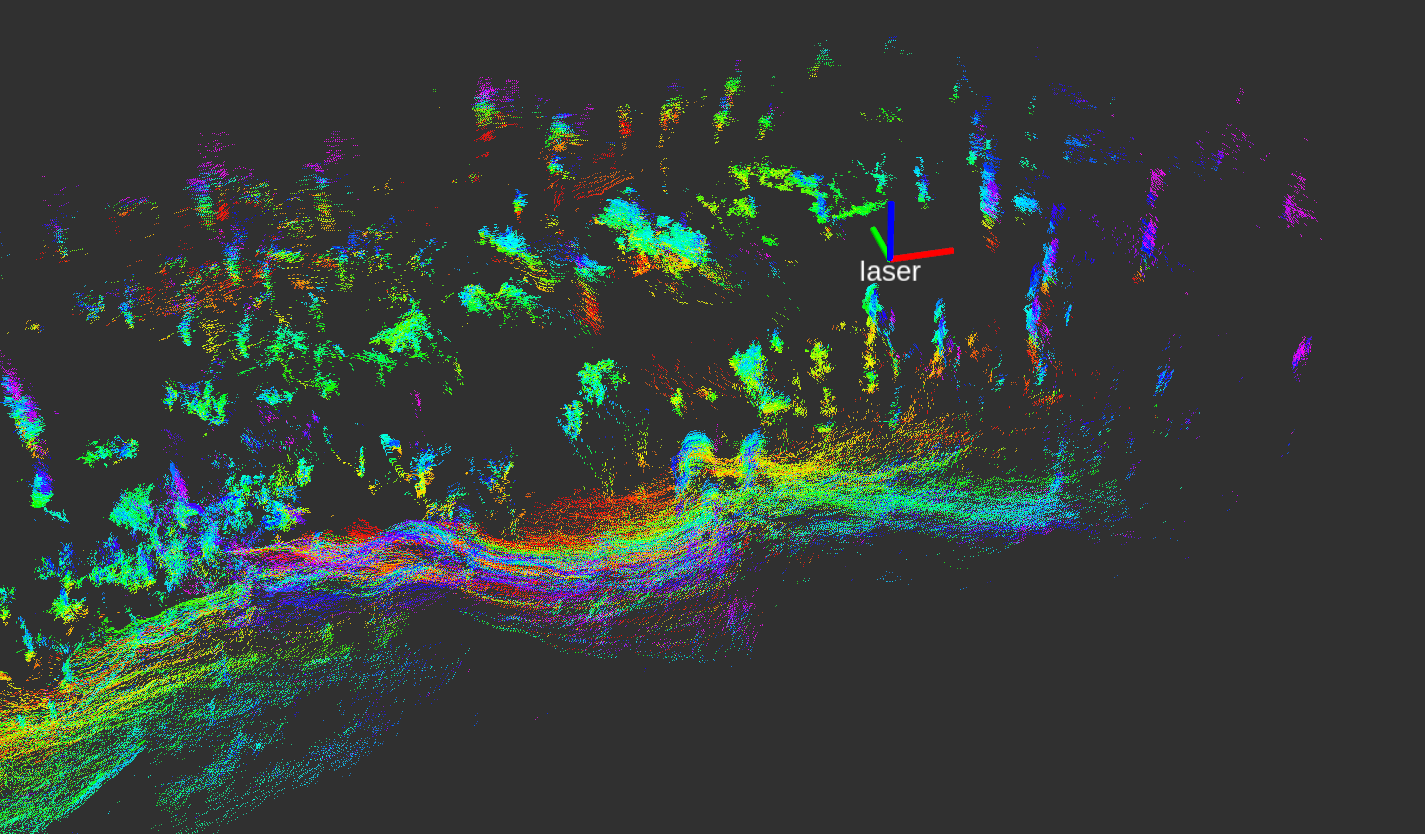}
	\caption{Raw point cloud collected by onboard laser scanner. Each individual point is colored according to its height from the ground. The vehicle pose estimated by the onboard EKF is represented as the \emph{laser} coordinate frame.}
	\label{fig:intro_laser}
	\end{subfigure}
	\caption{Autonomous flight of one of our UAVs in a forested area within NASA Langley Research Center (LaRC). A complete video of two UAVs autonomously exploring and collaboratively mapping the forest is available at: \protect\url{https://groups.csail.mit.edu/rrg/videos/iser2018}.}
	\label{fig:intro}
\end{figure}

Lost hikers are often within a mile or two of the last point of detection for extended
periods of time, but are undetected for hours because manned aircraft cannot see
them through the overhead forest canopy. 
Instead, a team of small autonomous unmanned aerial vehicles (UAVs) can be deployed \emph{under} the tree canopy to gain better visibility during search and rescue missions in forest areas;
these vehicles can
be rapidly deployed, can collaboratively explore expanses of terrain efficiently, and
are agile enough to operate in reasonably thick forests, such as shown in Figure~\ref{fig:intro}.

Deploying a multi-UAV system for forest search and rescue presents 
a multitude of technical challenges. 
Firstly, signals from the 
Global Positioning System (GPS) typically cannot penetrate the overhead forest canopy.
Thus, for independent and real-time operations,
vehicles must be able to perform GPS-denied flights
using onboard sensing, estimation, and planning. 
Secondly, a collaborative mapping system would require a map fusion procedure
that accurately detects overlaps between individual maps and fuses these maps into a globally consistent model of the environment. 
In the forests, however, correctly matching multiple overlapping maps
is a challenging problem due to
severe perceptual aliasing caused by the visual similarities of trees.
Lastly, communication constitutes another prerequisite for collaborative exploration and mapping.
However, in forest search and rescue scenarios, communication is typically unreliable and intermittent due to signal attenuation and limited transmission range. 

In this work, we present a multi-UAV system for efficient exploration in large-scale
unknown forest environments. Our vehicles are equipped with onboard autonomy that efficiently performs autonomous sensing, vehicle state estimation, local mapping, and exploration planning.
When communication is available, map fusion is achieved by performing collaborative simultaneous localization and mapping (CSLAM) at a central ground station.
We overcome the challenge of limited communication bandwidth by leveraging \emph{two different map representations}: a dense 3D volumetric grid representation for onboard trajectory planning, and a compact \emph{tree-based} representation for lightweight communication with the ground station during CSLAM.

Our onboard sensing, estimation, and mapping modules are based on a 2D LIDAR sensor, where laser scans are aligned frame-to-frame for incremental vehicle pose update, collected into a 3D occupancy grid for trajectory planning, and compressed into a tree-based map for lightweight communication.
We adopt a submap-based grid representation \citep{Leonard2001}, i.e., a collection of locally segmented grid maps as opposed to a single occupancy grid, which allows the origin of each submap in the world to be optimized to correct for onboard estimation drifts after loop closure. 
Our submapping approach also bounds the area that is updated after incorporating each new scan, but unlike the sliding-window mapping approach, e.g. by \cite{schmuckJFR2017}, all previous submaps are stored and used by a dynamics-aware frontier-based planner during exploration.
While dense local submaps are useful for onboard planning processes, na\"{i}ve transmission of rich geometric occupancy information poses high bandwidth requirements. 
For lightweight communication, we compress local maps into object-based maps by detecting tree stems in each laser scan.

On the ground station, we perform CSLAM to jointly optimize all submap origins, which determine how each submap should be positioned in the world frame, as well as tree positions in the fused map.
To do so, we first detect overlaps across individual submaps by performing loop closure detection. 
Instead of directly matching individual trees, which is prone to error due to their visual similarities, we match groups of trees using their relative positions and orientations as a unique signature. To gain additional robustness against perceptual aliasing, we include a novel stage during our data association 
process that jointly optimizes the set of all pairwise matches based on the criterion of \emph{cycle consistency} \citep{AraguesRSS2011}. Based on the final tree-to-tree correspondences between submaps, 
we perform landmark SLAM that jointly optimizes all submap origins and tree positions in the global map. This global map may then be inspected by the human operators at the ground station, and transmitted back to the vehicles to correct estimation drifts during onboard state estimation and mapping.

We demonstrate the proposed system in a small forest at NASA Langley Research Center (LaRC), where a team of two UAVs are each assigned a search region to completely explore (Figure~\ref{fig:intro}). 
\Yedit{
We show that the proposed loop closure detection and data association algorithms result in improved precision and recall when evaluated on datasets collected at the NASA LaRC forest. Using a high fidelity simulation environment, we also show that our dynamics-aware exploration planner outperforms the classical frontier-based planner in efficiently covering the search area, and that our SLAM pipeline effectively corrects onboard estimation drifts. }{}
Our experimental results show promising progress towards the practical use of multi-UAV systems for search and rescue under the forest canopy.

\Yedit{}{
\subsection{Contributions}

In this paper, we propose a novel collaborative mapping and exploration system that overcomes the unique challenges associated with forest search and rescue. 
It is widely recognized that data association (e.g., matching landmarks or feature points) is among the hardest problems for single and multi-robot SLAM. 
In a forest, data association is even harder due to large amount of noise caused by moving branches and leaves, and severe perceptual aliasing between different parts of the environment.
To overcome this problem, we utilize a novel tree-level map representation, and leverage both pairwise and \emph{multiway} matching techniques to improve the reliability of landmark association and guarantee \emph{cycle consistency}.
Using precision-recall as the performance crtical metric, we show that our approach significantly outperforms the baseline approach in a real-world forest environment.

In addition, for search and rescue, the efficiency to explore a designated search area is among the most crucial performance factors. 
Classical closest-frontier-based exploration strategy is inefficient for a UAV, as it requires frequent heading changes to visit frontiers that are behind the vehicle.
To address this issue, we utilize a heuristic for selecting frontiers that better respects vehicle dynamics.
Using a high fidelity simulation environment, we demonstrate the advantages of this approach over standard frontier-based exploration, in terms of both average vehicle speed and time efficiency.

}

An early version of this work was presented at ISER 2018
(International Symposium on Experimental Robotics) \citep{TianISER2018}.
This paper extends our previous work with the following contributions:
\begin{enumerate}
    \item Extension to a \emph{deformable} submap-based representation for local mapping (Section~\ref{sec:onboard_estimation}). 
    \Yedit{which allows for correction of local estimation drifts}
    {Using a high fidelity simulation environment, we show that the approach effectively corrects onboard estimation drift accumulated by EKF.}
    \item Novel use of \emph{cycle consistent multiway matching} \Yedit{during centralized data association for CSLAM}{for data association during multi-robot landmark SLAM} (Section~\ref{sec:clear}).
    \Yedit{}{We show that this approach further improves precision-recall performance compared to our previous work.}
    \item Additional experimental results that demonstrate the usefulness of the proposed multi-UAV exploration and mapping system (Section~\ref{sec:sim_experiments} and \ref{sec:real_experiments}).
\end{enumerate}

\subsection{Outline}
The rest of the sections are organized as follows. We first review previous work in Section \ref{sec:related} and give an overview of the proposed system in Section \ref{sec:overview}. We describe in detail the onboard autonomy in Section \ref{sec:onboard_autonomy} and the offboard CSLAM pipeline in Section \ref{sec:offboard_fusion}. We present simulation and real flight experiments and analyze the results in Sections \ref{sec:sim_experiments} and \ref{sec:real_experiments}. Finally, we share main experimental insights learned and suggest future work in Section \ref{sec:insights}.

\section{Related Work}
\label{sec:related}
\subsection{Multi-Robot Mapping and Exploration}
Multi-robot localization and mapping has been an active field of research. 
For a comprehensive survey,
see also
\cite{SaeediJFR2016,ChoudharyIJRR2017,schmuckJFR2017}
and the references therein.
\cite{ForsterIROS2013} are among the first to propose a 
centralized collaborative SLAM system for multiple UAVs. 
Each UAV runs visual odometry onboard and relays its keyframes and
relative pose estimates to a ground station. 
The ground station detects loop closures among local maps and
performs map fusion when necessary. 
\cite{schmuckJFR2017} extend this framework 
by sending 
optimized keyframes and map points back to the vehicles 
to improve accuracy of local mapping. 

On the other hand,
\cite{DongIRCA2015}, \cite{Morrison2016} and \cite{SchusterJFR2019}
propose to run full SLAM onboard each vehicle. 
The incurred computation costs are further reduced in \emph{distributed} architectures, 
where each robot only optimizes its local map and 
shares the compressed map or boundary poses with each other, 
see
\cite{CunninghamDDFSAM2,ChoudharyIJRR2017,CieslewskiICRA2018}.

While distributed systems have better scalability compared to centralized ones,
they do not leverage the opportunity to transfer expensive
onboard operations to the ground station which has more computational power
and is not limited by resource constraints. 
Furthermore, for search and rescue, 
a fused map ultimately has to be created on the ground station
and presented to the first responders. 
For these reasons, in this work
we choose to implement a centralized architecture similar to the ones 
presented in \cite{ForsterIROS2013,schmuckJFR2017}.

\Yedit{
The problems of search and exploration have been well-studied in the literature.
Standard approaches}
{The problem of multi-robot exploration have been well-studied in the literature.
Commonly used exploration strategies}
include nearest frontier selection \citep{YamauchiICRA1998},
next-best-view \citep{HectorIJRR2002,BircherICRA2016},
greedy or receding horizon information gain \citep{Feder99,Bourgault02},
or a mixture of these methods \citep{CharrowICRA2015}. 
\Yedit{More recently,
there have been results on exploring to find hidden objects in indoor environments \citep{aydemir2011search}.}
{More recently, exploration using a single or multiple UAVs has become increasingly popular \citep{Shen2012ICRA,CesareICRA15,Heng2015ICRA,Yoder2016FSR,CieslewskiIROS2017}.
Among these works, 
\cite{CesareICRA15} demonstrate 
a frontier-based exploration algorithm for multiple UAVs in an
indoor environment. 
To accelerate exploration, 
\cite{CieslewskiIROS2017} proposes to select
frontiers that are reachable with minimum change in velocity.
In this work, we draw similar intuitions and implement a \emph{dynamics-aware} strategy for exploration.
}

\Yedit{Multi-robot exploration has also gained considerable attention over the years.
Other related work for multi-robot exploration
focuses on deriving algorithms with provable performance guarantees \citep{Singh2009,CorahRSS2017}
and improving communication efficiency \citep{JensenIJCAI2018,CorahRAL2019}. }
{Another line of research focuses on coordinated multi-robot exploration.
\cite{Burgard2000ICRA,Burgard2005TRO} embed coordination into the exploration strategy by reducing the utilities of frontier cells that are seen by other robots. 
Experiments show that this approach results in more efficient exploration compared to the uncoordinated approach of \cite{YamauchiICRA1998}. 
The same approach is also adopted and extended by later works \citep{Fox2006,Stachniss2008}. 
This work differs with the aforementioned works in two aspects. 
First, earlier works frequently assume that loop closure can be performed straightforwardly, e.g., via scan matching.
In a forest, however, standard scan matching tends to fail due to large amount of noise caused by moving branches and leaves.
In this work, we propose and demonstrate a collaborative SLAM system that leverages prior structures within the environment to achieve reliable loop closure and map fusion.
Second, \citep{Burgard2005TRO,Stachniss2008} also rely on a central hub to assign frontiers to robots. 
In practice, this centralized approach sometimes create unnecessary latency in the system
(e.g., a vehicle could become idle while waiting for the next frontier from the ground station).
We overcome this problem by implementing a distributed exploration strategy where each vehicle independently explores a non-overlapping search region. 
}

\subsection{Multi-Robot Loop Closure and Data Association}
\Yedit
{Robust loop closure detection and data association form the backbone
of any CSLAM system.
Existing approaches for multi-robot loop closure and data association
can be categorized into
\emph{pairwise} and \emph{multiway} techniques.}
{One major contribution of this work is the design of a reliable multi-robot loop closure and map fusion system for the forest environment. 
This section review existing techniques on multi-robot loop closure and data association, 
which can be largely categorized into \emph{pairwise} and \emph{multiway} methods.}

\subsubsection{Pairwise Methods} 
Detecting correspondences between 
two sets of feature points is a well-studied problem in robotics. 
Classical \emph{geometry-based} methods such as  
Nearest Neighbor, Maximum Likelihood, and Joint Compatibility Branch and Bound (JCBB) 
associate pairs of landmarks based on their Euclidean or 
Mahalanobis distances in a common reference frame \citep{NeiraTRA2001,ZhouIROS2006,KaessRAS2009,GilRAS2010}. 
When a common reference frame is not available, methods based on local geometric features \citep{CunninghamICRA2012},
Generalized Hough Transform \citep{PazIROS2005},
or Maximum Common Subgraph (MCS) \citep{BaileyICRA2000} can be used instead.
Many of the above techniques can be combined with RANSAC \citep{FischlerRANSAC}
for enhanced robustness. 

Recently, \emph{appearance-based} methods have shown great success in
large scale place recognition \citep{PaulICRA10,CumminsIJRR11}
and distributed visual SLAM \citep{CieslewskiICRA2018}.
Many of these methods start by detecting local features from input images or laser scans.
Each feature point is typically associated with a descriptor.
For visual data, 
popular descriptors include
SIFT \citep{LoweSIFT2004}, SURF \citep{BaySURF2006}, BRIEF \citep{CalonderBRIEF2010}, 
and ORB \citep{RubleeORB2011}.
For laser-range data, similar descriptors have been developed 
which include Gestalt \citep{BosseRAS2009}, FPFH \citep{RusuICRA2009}, and FLIRT \citep{TipaldiICRA10}.

Although feature extraction is very fast, 
the number of features needed for tracking and loop closure detection may still pose a challenge for
real-time computation and communication. 
To address this issue, recent works begin to use more lightweight
representations based on \emph{semantic} and \emph{object-level} models.
\cite{ChoudharyIJRR2017} use objects detected by convolutional neural networks
during distributed pose graph optimization. \cite{SegmapRSS2018}
represent objects with segments in 3D point clouds and use data-driven descriptors 
for loop closure detection. 

In the forest environment, prior works have
recognized trees as the distinctive objects that can aid localization and mapping
\citep{AsmarCVPRW2006, OhmanFSR2008, TangForests2015, KukkoISPRS2017}. 
\Yedit{The present work also uses trees as landmarks in the map representation. However,
unlike prior works that consider passive SLAM with a single robot, 
we focus on the more general scenario of collaborative SLAM 
during multi-robot autonomous exploration.}
{In this work, we also use trees as informative features for data association and loop closure.
However, unlike prior works such as \citep{KukkoISPRS2017}, we develop on top of tree features a complete data association procedure that leverages both pairwise and cycle consistent multiway matching.
In our experiments, we demonstrate our system in a real-world multi-robot exploration task beneath the forest canopy.}

In addition,
state-of-the-art SLAM systems often implement a place recognition module
that aggregates local features 
into a global descriptor for fast detection of potential loop closures. 
For images, notable examples include bag of visual words \citep{LopezTRO12},
vector of locally aggregated descriptors (VLAD) \citep{JegouCVPR2010},
Fisher vectors \citep{JegouFisher2012,PerronninCVPR2010}, 
and deep learning based methods such as NetVLAD \citep{ArandjelovicCVPR2016}. 
For laser scans, \cite{HimstedtIROS14} develop the 
Geometric LAndmark RElations (GLARE) signature, which is later extended by
\cite{KallasiIROS16} to be rotationally invariant in GLAROT. 
Note that comparing global descriptors only yields a set of candidate matches. State-of-the-art SLAM systems still employ a \emph{geometric verification} step to validate the proposed associations; 
see e.g., \cite{MurTRO2017}.

\subsubsection{Multiway Methods}
For robust data association in robotics, multiway association that  simultaneously considers multiple noisy pairwise associations is often used. One prominent example is the problem of rejecting outlier loop closures in pose graph SLAM \citep{IndelmanICRA2014,DongIRCA2015,JoshuaICRA2018}.
Among the different ways of performing multiway association, a family of techniques closely related to our work attempts to achieve \emph{cycle consistent}
multi-robot data associations \citep{AraguesRSS2011,MontijanoTRO2013}.
Specifically, these works seek to resolve inconsistencies in the data associations; for example, such inconsistencies can appear when a chain of spurious pairwise associations matches two landmarks observed by the same robot. 
\cite{AraguesRSS2011} propose a heuristic based on cycle detection
to detect and resolve these inconsistencies. 

Similarly in the computer vision community,
cycle consistency has also gained considerable attention due to popular applications such as multi-shape matching \citep{NguyenCGF2011,HuangSGP2013}
and multi-image matching \citep{ZhouICCV2015,LeonardosICRA2017}. 
Principled approaches based on 
spectral relaxation \citep{PachauriNIPS2013},
semidefinite relaxation \citep{ChenICML2014}, 
distributed consensus \citep{LeonardosICRA2017},
and spectral clustering \citep{FathianCLEAR2019} have been proposed for solving this problem, 
and performance guarantees for exact matching are proved under certain noise models \citep{PachauriNIPS2013,ChenICML2014}.
In this work,
\Yedit{
we leverage these recent advances to achieve cycle consistent data associations
in our robust collaborative SLAM.}
{we propose a novel application of cycle consistent multiway matching for fusing tree landmarks during multi-robot data association and SLAM.}

\begin{figure*}[t]
	\centering
    \begin{subfigure}{0.48\textwidth}
	\includegraphics[width=1.0\textwidth]{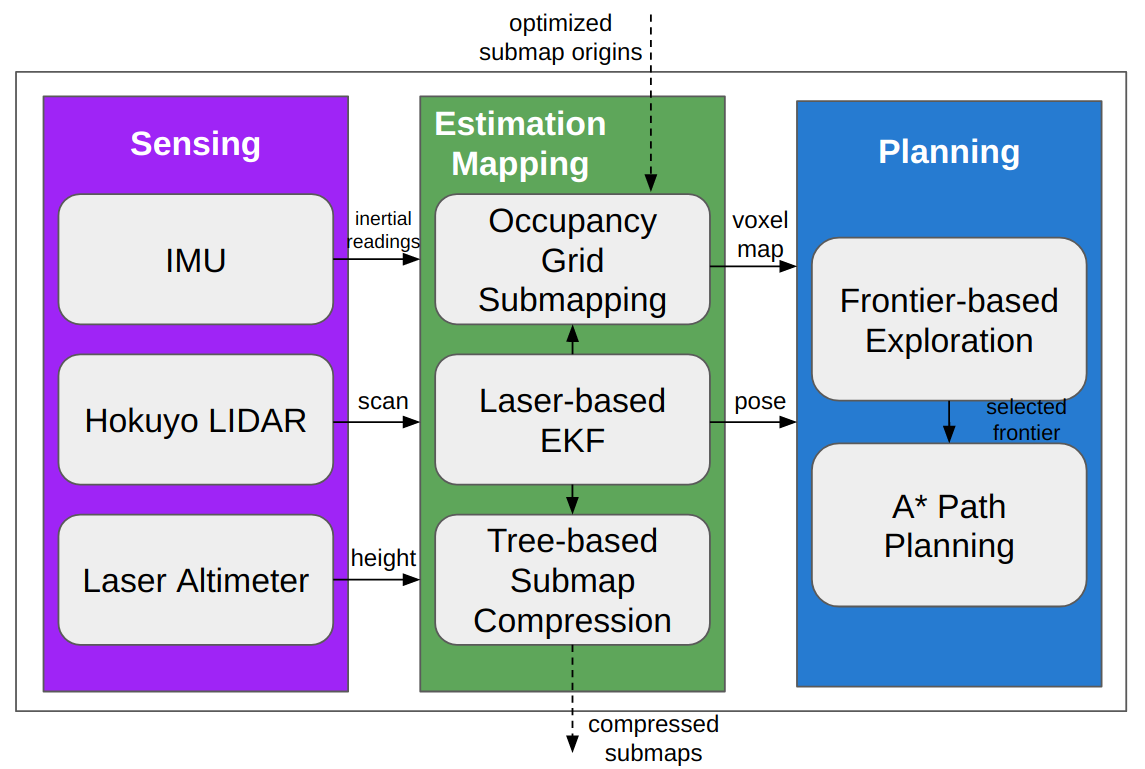}
	\caption{Onboard Autonomy}
	\label{fig:onboard_diagram}
	\end{subfigure}
	~
	\begin{subfigure}{0.48\textwidth}
	\includegraphics[width=1.0\textwidth]{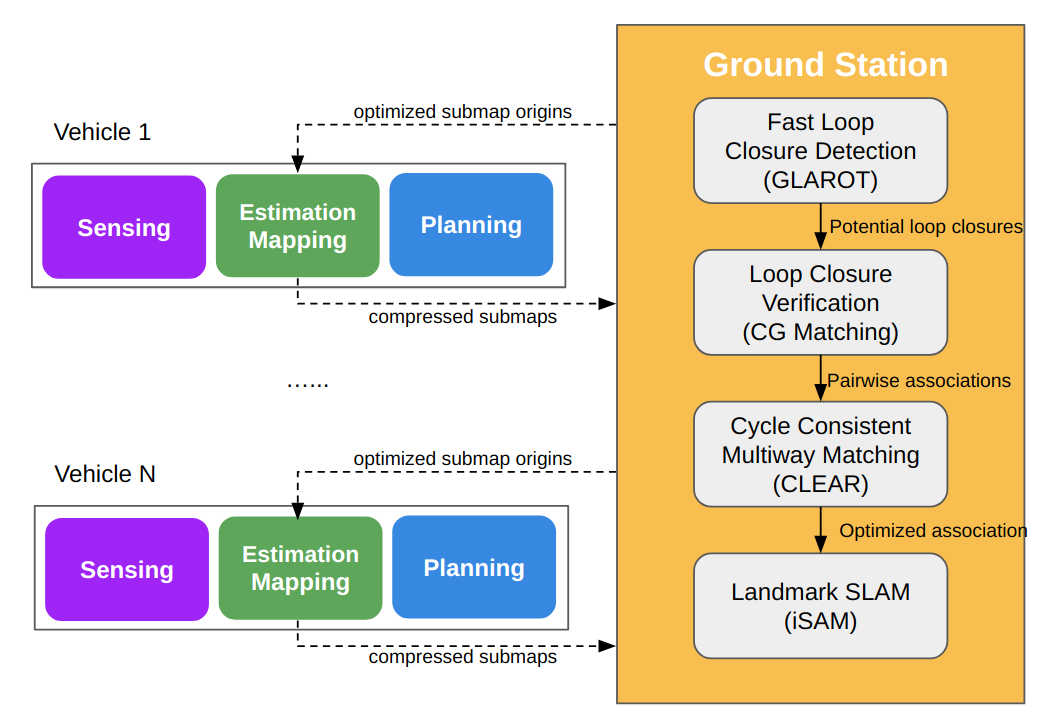}
	\caption{Centralized Offboard Mapping}
    \label{fig:offboard_diagram}
	\end{subfigure}
	\caption{System architecture. Dashed lines denote communication.
	\textbf{(a)}:
	Each vehicle achieves full autonomy by performing onboard sensing, vehicle state estimation, local mapping, and exploration planning.
	\textbf{(b)}:
    Ground station performs CSLAM by detecting matches (loop closures) between received submaps, recovering associations between trees observed in multiple submaps, and optimize all submap origins and tree positions in the fused map. }
	\label{fig:system_diagram}
\end{figure*}

\section{Overview}
\label{sec:overview}
To enable independent exploration of large-scale GPS-denied forest environments,
we equip each of our UAVs with an onboard autonomy module that performs 
autonomous sensing, vehicle state estimation, local mapping, and exploration planning;
see Figure~\ref{fig:onboard_diagram}.
We adopt a filtering-based approach to infer the vehicle state, while 
merging incoming laser scans into an occupancy grid 
for real-time obstacle avoidance and exploration planning.
When communication is available, each vehicle also transmits 
its local observations to a central ground station to contribute to the collaborative mapping process.
To cope with the limited communication bandwidth, 
these local observations are first compressed into
lightweight submaps consisting only of detected tree objects in the environment.
A detailed description of each component of the onboard autonomy module is provided in Section~\ref{sec:onboard_autonomy}.

The ground station performs CSLAM which fuses noisy measurements (in the form of tree-based submaps) from different vehicles
into a globally consistent map of the world.
This is achieved via a two stage process that consists of 
(i) \emph{global data association}, which 
recovers the underlying correspondences between trees observed across multiple submaps and possibly by different vehicles,
and (ii) \emph{landmark SLAM},
which jointly optimizes all submap origins and tree positions in the global map. 
See Figure~\ref{fig:offboard_diagram} for the corresponding system diagram.
The optimized submap origins are transmitted back to the vehicles to correct estimation drifts during local mapping.
We note that the offboard CSLAM
can happen in parallel to the onboard autonomous operations of each vehicle.
This centralized architecture provides 
sufficient tolerance against network failures which in our architecture would only delay potential map fusion,
but will not hinder real-time onboard operations.
We describe the centralized CSLAM pipeline in detail in Section~\ref{sec:offboard_fusion}.

\section{Onboard Autonomy}
\label{sec:onboard_autonomy}
Each of our vehicles achieves full onboard autonomy by estimating the vehicle pose in real-time and simultaneously optimizing over a 
local geometric map for obstacle avoidance and exploration planning.
The rest of this section is organized as follows.
We first describe our approach for laser-based vehicle pose estimation and 3D occupancy grid mapping (Section~\ref{sec:onboard_estimation}). 
We then describe our frontier-based exploration planning algorithm (Section~\ref{sec:frontier}), 
and lastly our object-based map compression scheme
suitable for low-bandwidth communication in the forest (Section~\ref{sec:compression}).

\subsection{State Estimation and Local Mapping}
\label{sec:onboard_estimation}
Forests are challenging environments for lightweight onboard sensing.
Vision-based sensors, such as RGB cameras that provide rich information about the environment, suffer from occlusions between trees, large lighting changes with movements of foliage in the wind, visual similarities between trees, and limited field of views.

Following \cite{GiamouIROS2017}, we instead adopt a 2D Hokuyo LIDAR as our main onboard sensor
as it provides a large field of view ($270^\circ$), accurate depth information, and robustness to visual aliasing and lighting changes,
and is also lightweight enough to be carried by a smaller UAV.
In addition to the LIDAR, we also utilize a single-point downward-facing laser altimeter and an inertial measurement unit (IMU),
both of which provide additional information about vehicle state.

We combine the raw measurements produced by the suite of onboard sensors in a laser-based Extended Kalman Filter (EKF) 
\citep{bachrach2011range} that estimates the 6 degrees-of-freedom pose of the vehicle. We first estimate the incremental motion between two consecutive laser scans using the iterative closest point (ICP) algorithm, and subsequently fuse this estimate with height measurements from the laser-based altimeter and inertial measurements produced by the IMU.
\Yedit{}{ 
Due to dense geometric information provided by laser scans as well as the close simiarlity between consecutive scans, ICP in practice often provides accurate estimate of incremental vehicle motion. 
Furthermore, ICP is also computationally efficient, which makes it favorable for high-rate state estimation.
In our implementation, we run ICP onboard at the same rate as raw laser scan comes in ($40$ Hz). }

Each vehicle uses the real-time state estimates from EKF to build a map of the explored region. In our previous work \citep{TianISER2018}, each onboard map was a 3D occupancy grid produced using Octomap \citep{hornung13auro}. While occupancy grid is a popular map representation especially for the purpose of path planning, one of its drawbacks is that it is not easily amenable to correction of previous mapping errors. In our case, such corrections are often needed to account for onboard estimation drifts, which is important for the vehicle to have a more accurate understanding of the environment (e.g., in terms of explored and unknown space) and adjust its decision intelligently to maximize search efficiency. 

In this work, we extend our onboard map representation to a \emph{submap-based} representation \citep{Leonard2001}. Each onboard map is stored as a collection of smaller 3D occupancy grids.
This approach makes the onboard map \emph{deformable}: while each submap is locally rigid, the relative transformations between submaps can be optimized using the refined pose estimates provided by CSLAM. This allows us to correct the errors accumulated during onboard vehicle estimation. An additional benefit is that at any time, the vehicle only needs to update its current submap which is computationally more efficient.

In our implementation, we initialize a new empty submap (Octomap) on each vehicle after a fixed amount of time.\footnote{Alternatively, one can initialize a new submap after the vehicle travels a fixed distance or after the estimation uncertainty of EKF grows beyond a threshold.} To update the current submap, we transform each incoming laser scan from the local sensor frame to the coordinate frame of the submap. The resulting point cloud is then used to update the 3D octomap. Although each submap only provides local information of a small area, the union of submaps contains the same amount of information as a single global map. Therefore, while we only update the current submap during onboard mapping, we use the union of the submaps for onboard motion and exploration planning.

Several state-of-the-art systems, e.g., \citep{schmuckJFR2017}, use a sliding window approach for local mapping to limit the onboard memory usage. 
However, such approach that discards old information is not suitable for our vehicles because they must keep track of all areas previously visited during exploration. In addition, we note that in practice the memory usage per vehicle is relatively small as each vehicle is tasked with exploring a reasonably bounded search region.

\subsection{Frontier-based Planning}
\label{sec:frontier}
We adopt a frontier-based algorithm for onboard exploration, where a \emph{frontier} refers to a location in the current map that is on the boundary of known and unknown space. 
\Yedit{Our choice is based on the low computational cost of frontier-based algorithms 
compared to algorithms based on maximizing information gain.}
{We select frontier-based exploration as it is computationally lightweight, which makes it easy to run distributedly onboard each vehicle.}
After the frontier-based planner selects the next frontier to visit,
a motion planner based on A* search \citep{Astar} is used on the projected 2D occupancy map 
to plan the optimal collision-free path for the vehicle. 
Exploration finishes when there are no more frontiers
left in a designated search area.

While frontier-based algorithms are computationally efficient and thus more suitable for lightweight platforms, finding the best frontier for exploration using a UAV can be a challenging problem.
Classic approaches usually select the closest frontier based on the Euclidean distance \citep{YamauchiICRA1998}.
However, when using maps built with our onboard mapping system, 
it is often the case that the closest frontier lies directly behind the vehicle,
in the blindspot of the $270^\circ$ field-of-view LIDAR. 
Selecting these frontiers would yield rapidly changing vehicle orientations that ultimately lead to a poor mapping performance and a lack of progress towards searching uncovered ground.

In this work, we select frontiers based on a hybrid cost function
that accounts for both the distance to frontiers and the vehicle dynamics,
similar to the technique in \citep{CieslewskiIROS2017}.
We define the cost associated to a frontier $a$  as
\begin{equation}
    J(a) = J_\theta(a) + \lambda J_t(a),
\end{equation}
where $J_t(\cdot)$ is the classic Euclidean distance of $a$, 
$J_\theta(\cdot)$ is the change in orientation required to reach $a$ from the vehicle's current heading,
and $\lambda$ is a balance parameter that trades off the two cost terms.
\Yedit{}
{In practice, the choice of $\lambda$ depends heavily on the flight speed. 
At lower speed, our vehicle can handle more aggressive heading change, which means that $\lambda$ can be increased to place more weight on the distance to frontier.
In our outdoor experiments, we have commanded both vehicles to fly at a constant speed of $2.0$ m/sec.  
At this speed, we observed that our choice of $\lambda = 0.5$ leads to the best empirical performance.}

Directly incorporating the orientation change in the cost function discourages the vehicle from excessive turning and hence produces a smoother overall trajectory.
In addition, accounting for orientation change also 
enables \emph{continuous} update of the next-best frontier, i.e., at any time during flight, 
the planner can replace the currently selected frontier with a new frontier whose cost is significantly lower.
Because of the orientation cost term $J_\theta(\cdot)$,
switching to the new frontier typically produces minimal heading changes and thus encourages safe flight behavior.
In contrast, a dynamics-agnostic planner, e.g., with a purely distance-based cost, has to wait 
until the vehicle reaches the current frontier to be able to select the next frontier, 
which is not time efficient. 

\subsection{Tree-based Map Compression}
\label{sec:compression}

\begin{figure}[t]
\centering
\begin{subfigure}{0.33\textwidth}
\includegraphics[width=1\textwidth,trim={0cm 0cm 0cm 0cm},clip]{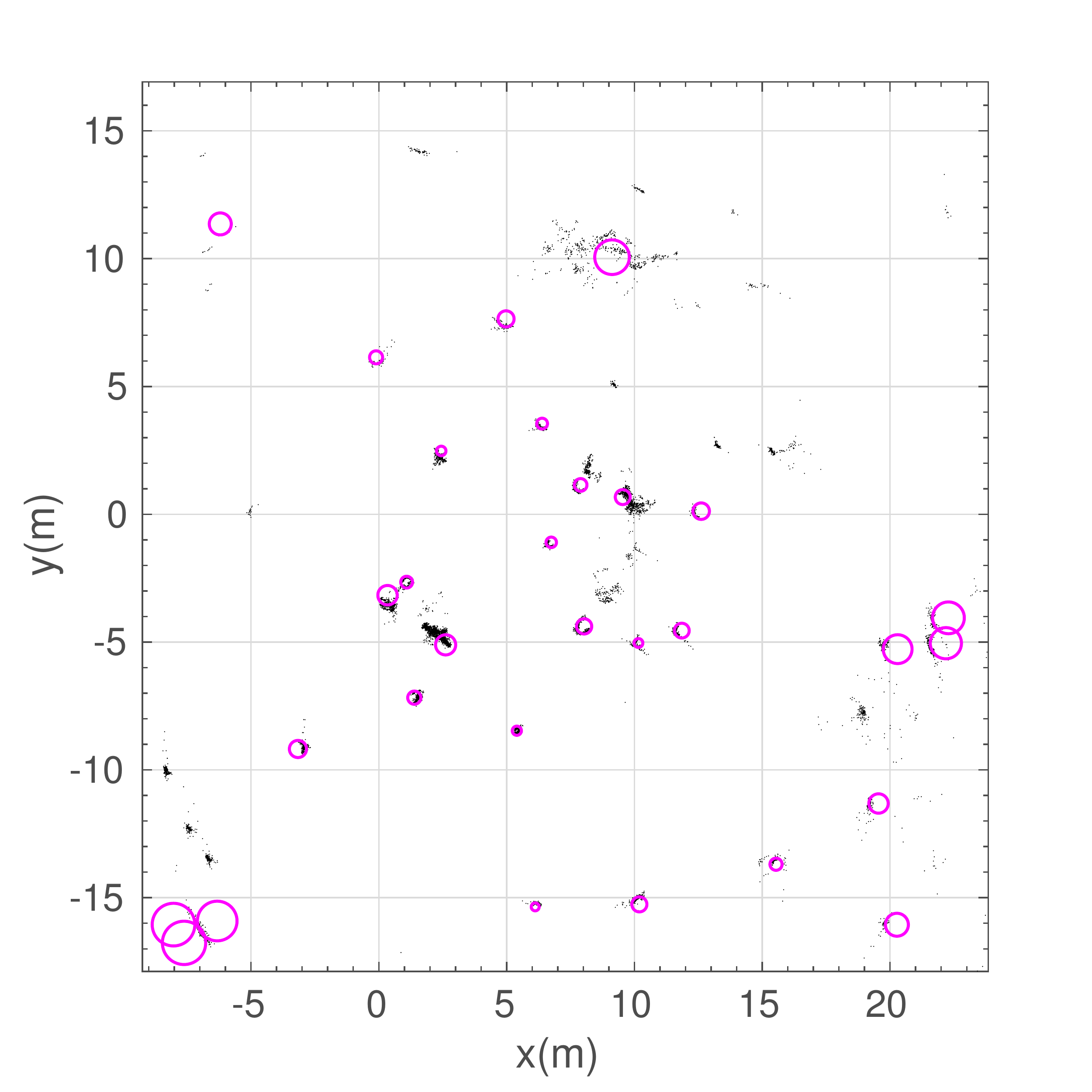}
\caption{Tree submap}
\label{fig:tree_submap}
\end{subfigure}
~
\begin{subfigure}{0.12\textwidth}
\includegraphics[width=1\textwidth,trim={0cm 0cm 0cm 0cm},clip]{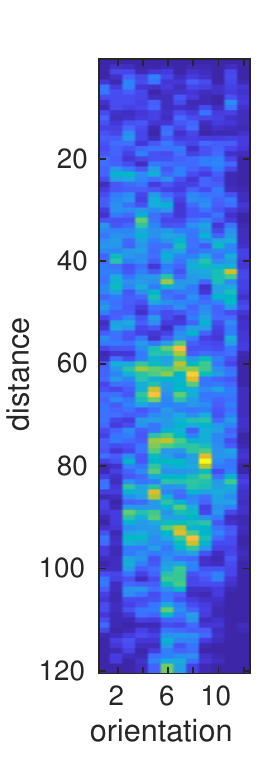}
\caption{GLARE signature}
\label{fig:glare}
\end{subfigure}
\caption{
\textbf{(a)}:
Example submap generated from real forest 
data collected at NASA LaRC. 
Black dots show the raw laser returns over a period of $20$ seconds.
Each magenta circle shows a tree tracked within this submap,
where the radius corresponds to the estimated 
tree stem radius.
\textbf{(b)}:
The corresponding GLARE signature. Horizontal and vertical axes show discretizations with respect to orientation and distance, respectively.
Lighter color denotes higher value.
}
\label{fig:submap}
\end{figure}

While dense volumetric maps provide accurate geometric information for path planning, they are not data efficient
and hence are not suitable for communication over a low-bandwidth wireless network.
Other representations such as 3D point clouds \citep{SchusterJFR2019} or feature-based maps \citep{schmuckJFR2017} can be more lightweight, but both can still result in potentially heavy data payload.

To achieve lightweight communication,
we turn to \emph{object-based} representations, more specifically a \emph{tree-based} representation \citep{KukkoISPRS2017}. 
Before transmission, we compress each submap into a sparse collection of trees as shown in Figure \ref{fig:tree_submap}.
This is done by first clustering beams in each LIDAR scan using Dirichlet process clustering (DP-means) \citep{CampbellNIPS13}.
\Yedit{Each cluster of points is then fitted to a circle in 2D that represents a tree trunk, }
{Each cluster of points is first projected to the 2D plane, and then fitted to a circle that represents a tree trunk.}
\Yedit{by applying}
{We first apply}
an algebraic fitting algorithm  \citep{Taubin1991}
that estimates a tree position and radius and returns a residual error.
If the initial residual error after the algebraic fitting is sufficiently low,
a geometric fitting process based on Levenberg-Marquardt \citep{Chernov2010}
that uses the algebraic fitting as initial guess is employed
to refine the estimated tree parameters.
A tree detection is accepted if the final residual error after geometric fitting is less than
$0.015$, the tree has a radius greater than $0.1$ m, and the observed LIDAR beams
cover more than 30\% of the tree trunk.

For improved stability, 
we also perform tree tracking and culling within each submap.
At every time step, a newly detected tree is combined with the closest
previously detected tree in the same submap
if their estimated positions are within $0.5$ m
and their radii differ less than $10\%$. 
If the above two conditions hold, the observation count of the original tree is incremented by one; if not, a new tree is initalized in the current submap using parameters of the newly
detected tree.
Before transmitting to the ground station, 
we optionally perform {culling} inside the submap,
by removing trees with observation counts less than a threshold
$\tau_\text{cull}$.
Here, $\tau_\text{cull}$ is a tunable parameter
designed to cope with varying level of noise 
during tree detection.

\Yedit{}
{
Our current tree detection algorithm assumes that trees can be represented as cylinders in 3D (or circles after projection to 2D). 
In practice, we find that this approximation works well if the vehicle is flying at an altitude at which most tree trunks can be seen clearly. 
On the other hand, performance of this method will degrade for denser forests with more branches and leaves. 
In these scenarios, more sophisticated detection algorithms are needed to improve robustness. 
}

\section{Offboard Collaborative SLAM}
\label{sec:offboard_fusion}

On the ground station, we construct a globally consistent map of the world by fusing identical trees detected across multiple submaps
and performing landmark SLAM over the fused trees.
The overall problem is combinatorial in nature 
and thus difficult to solve  optimally;
however, practical applications, e.g., search and rescue, often require a solution that is efficient and, ideally, real-time.
In order to meet such performance demand,
we make use of the following centralized pipeline to efficiently (albeit approximately) 
solve the global mapping problem on the ground station; see Figure~\ref{fig:offboard_diagram}.
First, we identify {potential} matches between pairs of submaps, i.e., potential loop closures, using a compact descriptor developed for 2D point landmarks (Section~\ref{sec:glare}). 
We then verify each potential match
by solving for the pairwise correspondences between trees in the two submaps (Section~\ref{sec:cg}).
Finally, the set of all pairwise associations are verified based on cycle consistency (Section~\ref{sec:clear}) to overcome perceptual aliasing.
Given the final data association results, we perform landmark SLAM
to jointly optimize all submap origins and tree posistions in the global map (Section~\ref{sec:isam}).

\subsection{Fast Loop Closure Candidate Detection}
\label{sec:glare}

To efficiently detect potential loop closures between pairs of submaps, 
we build a GLARE \citep{HimstedtIROS14} descriptor for each received submap on the ground station. 
GLARE is an efficient method for encoding relative landmark (i.e. tree) positions in a single map.
For every pair of trees $(i,j)$, we compute their distance 
$\rho_{i,j}$ and absolute relative angle $\theta^+_{i,j}$.
We then assign these values to bins in a 2D histogram $(\theta^+_{i,j}, \rho^t_{i,j})\in
\textrm{bin}(n_\theta,n_\rho)$,
where $n_\theta \in \{1,2,\hdots,N_\theta\}$ and $n_\rho \in \{1,2,\hdots,N_\rho\}$ 
are indices corresponding to the quantization
of $(\theta^+_{i,j}, \rho^t_{i,j})$
in the range $[0, \rho_{\max}]$m $\times$ $[0, \pi]$rad.
To account for noise during tree detection, we apply a
Gaussian blur \Yedit{}{with standard deviation 0.1} to the histogram of each feature pair.
The GLARE descriptor is computed as the sum of the 2D histograms over all pairs.
Following \cite{GiamouIROS2017}, we set the default histogram resolutions to be $N_\rho = 120, N_\theta = 12$.
\Yedit{}
{In our experiments, we have not observed significant impact of the discretization parameters on the performance of loop closure detection.}
Figure~\ref{fig:glare} shows the GLARE signature corresponding to the submap in Figure~\ref{fig:tree_submap}.

To determine if two submaps are likely to be a loop closure,
we compare their GLARE descriptors using the GLAROT procedure proposed by \cite{KallasiIROS16}.
GLAROT provides a rotation-invariant distance metric, by computing the {shifted L$_1$ distance} between two GLARE descriptors $G^s, G^t \in \mathbb{R}^{N_\rho \times N_\theta}$, 
\begin{equation}
\small
SL_1(G^s, G^t) = \min_{0 \leq k < N_\theta} \sum_{i=0}^{N_\rho - 1} \sum_{j=0}^{N_\theta - 1} |G^t_{i,j} - G^s_{i,(j+k) \text{ mod } N_\theta}|.
\end{equation}
Each GLAROT query requires $N_\rho \times N_\theta^2$ operations which is very fast on standard CPUs.
A pair of submaps $(M^s, M^t)$ is declared to be a loop closure candidate, if their GLAROT distance
$SL_1(G^s, G^t)$
is below a threshold $\epsilon_\text{GLAROT}$.
This process serves as an efficient filtering step that reduces the search space for loop closures to a small set of candidate matches.
To verify if each candidate is a true loop closure, 
we apply a pairwise data association procedure
which identifies the tree-to-tree correspondences between the two submaps.

\subsection{Loop Closure Verification}
\label{sec:cg}
For each potential loop closure between a pair of submaps $(M^s, M^t)$, we verify if $(M^s, M^t)$ forms a true loop closure
by solving an element-wise data association problem that returns the correspondences between trees in the two submaps.
Note that as $M^s$ and $M^t$ could potentially come from different vehicles, 
classical pairwise association methods such as Nearest Neighbor or JCBB which rely on a common reference frame 
could not be used. Instead, we perform correspondence graph (CG) matching \citep{BaileyICRA2000},
which is observed to work better 
compared to alternative methods based on the Generalized Hough Transform 
or RANSAC \citep{GiamouIROS2017}.
Let $\{o^s_i\}$ and $\{o^t_j\}$ represent the set of objects in $M^s$ and $M^t$.
In addition, 
let $\{p^s_i\}, \{p^t_j\}$ 
be the positions of these objects
in the coordinate frames of $M^s$ and $M^t$, respectively.
A correspondence graph is an undirected graph in which each 
vertex $u^{s,t}_{i,j}$ denotes a hypothetical match between 
$o^s_i \in M^s$ and $o^t_j \in M^t$.
Two vertices $u^{s,t}_{i,j}, u^{s,t}_{k,l}$ are connected by an edge (i.e., they are geometrically consistent) if relative distances between landmarks are preserved, i.e., if,
\begin{equation}
    |\norm{p^s_i-p^s_k}_2 - \norm{p^t_j - p^t_l}_2| \leq \epsilon_\text{\tiny CG},
\end{equation}
where $\epsilon_\text{\tiny CG}$ is an adjustable tolerance parameter to account for noise in real measurements.

Given the correspondence graph, we find the maximal set of pairwise compatible correspondences between $M^s$ and $M^t$ by finding the maximum clique. 
The resulting data association can be written as a 
partial permutation $\tilde{\pi}^s_t$ which maps a subset of objects in $M^s$ to $M^t$.
We use the tilde notation as a reminder that
$\tilde{\pi}^s_t$
could still contain wrong associations, e.g., due to strong perceptual aliasing in the forest.
For numerical computations in the next section,
each $\tilde{\pi}^s_t$ is also represented as 
a partial permutation matrix $\tilde{\Pi}^s_t \in \{0,1\}^{m^s \times m^t}$, where
\begin{equation}
    \tilde{\Pi}^s_t(i,j) = \begin{cases}
    1, \text{ if }\tilde{\pi}^s_t(i) = j, \\
    0, \text{ otherwise},
    \end{cases}
\end{equation}
and $m^s$ and $m^t$ are the sizes of the two submaps.
$M^s$ and $M^t$ are declared to be a loop closure,
if $\tilde{\pi}^s_t$ contains sufficient amount of matches,
specified by a tunable parameter $\tau_\text{\tiny CG}$.

We make a note on the complexity of CG matching.
Although the maximum clique problem is combinatorial in nature,
our object-based representation allows us to
solve it efficiently and in real time.
To solve the maximum clique problem, we use the implementation 
provided by \cite{Konc2007}.
Empirical runtime results are reported in Section~\ref{sec:real_experiments}.

\subsection{Globally Consistent Loop Closures}
\label{sec:clear}

\begin{figure*}[t]
\centering
\begin{subfigure}{0.30\textwidth}
\includegraphics[width=1\textwidth]{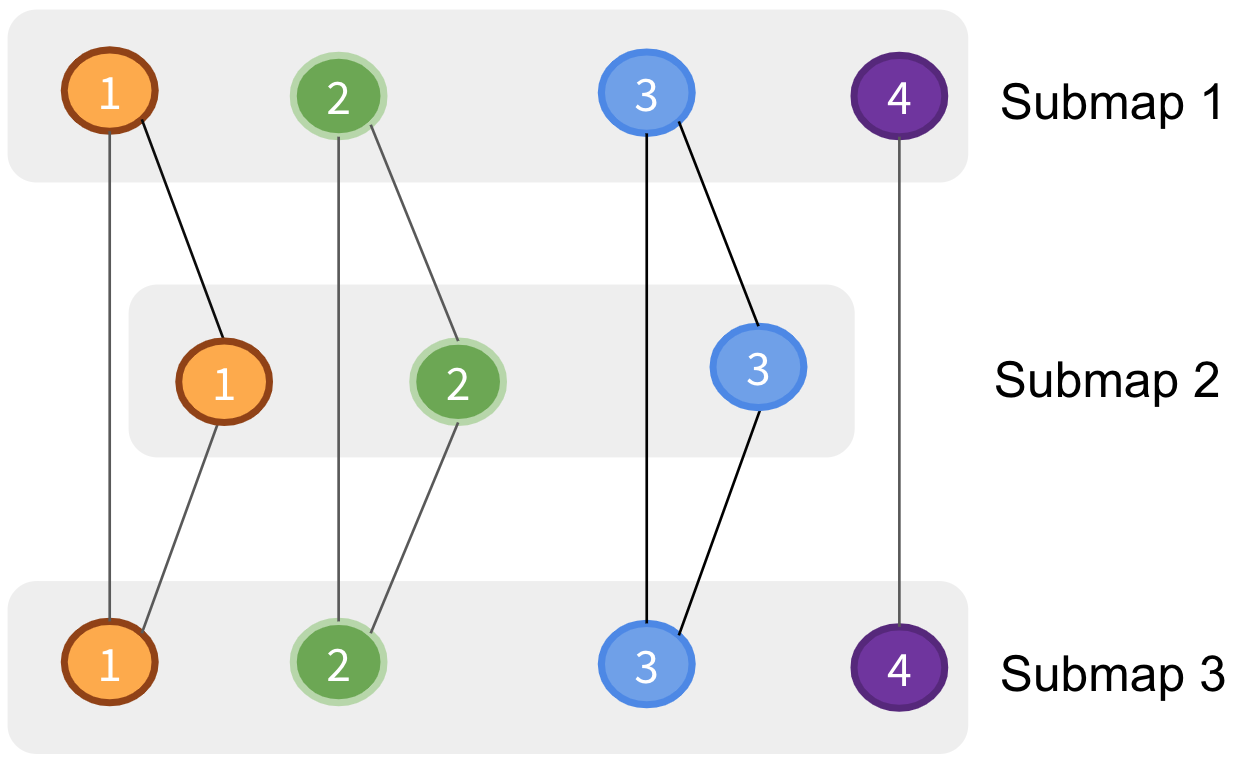}
\caption{Cycle consistent induced graph}
\label{fig:consistent_induced_graph}
\end{subfigure}
\hspace{2cm}
\begin{subfigure}{0.30\textwidth}
\includegraphics[width=1\textwidth]{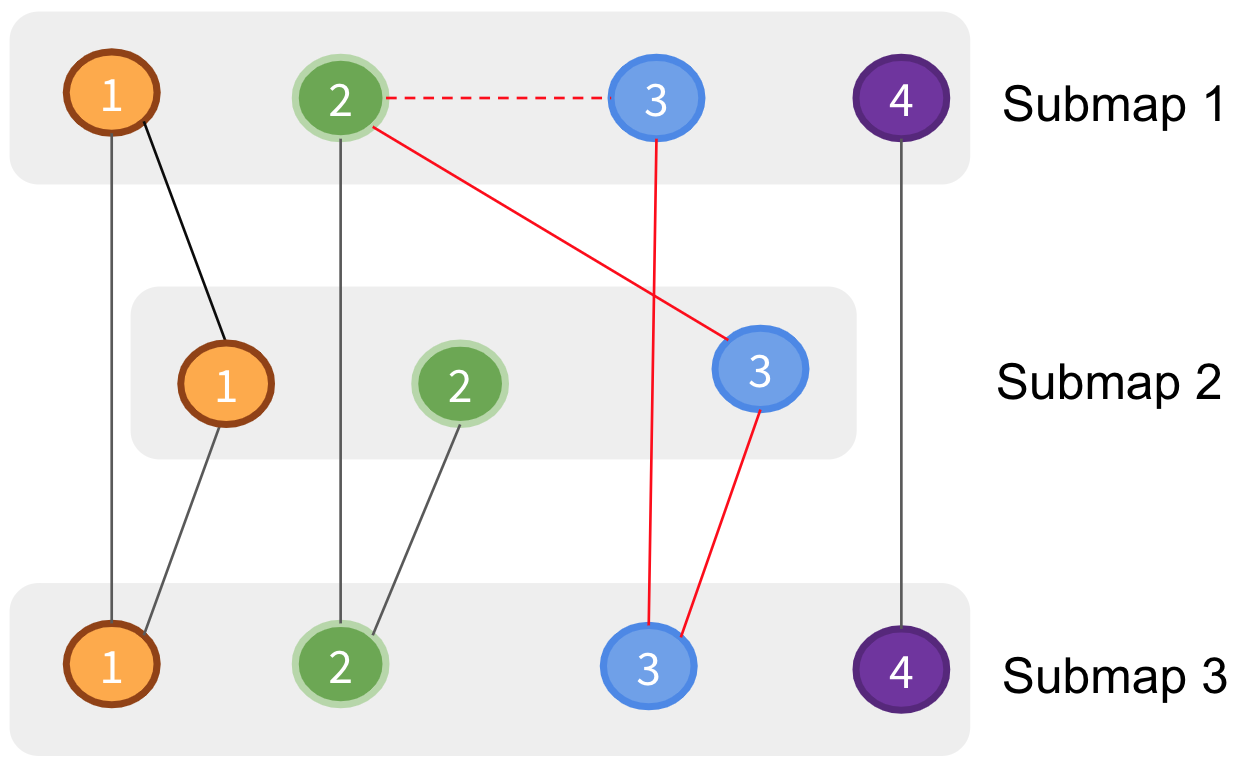}
\caption{Inconsistent induced graph}
\label{fig:inconsistent_induced_graph}
\end{subfigure}
\caption{
Example data association graphs.
Each vertex (circle) represents an object (tree) in a submap.
Two vertices are connected if the corresponding objects
are matched in the input pairwise associations.
\textbf{(a)}:
If the input associations are cycle consistent,
objects matched together will form disjoint cliques.
In this case, it is straightforward to read out the global data association by assigning each clique to an object in the universe.
\textbf{(b)}:
In practice, a chain of spurious associations (red edges) 
might violate cycle consistency and
incorrectly fuse two objects in the same submap (red dashed edge).
We remove such noisy associations based on cycle consistency.
}
\label{fig:induced_graphs}
\end{figure*}

Given a collection of noisy pairwise associations $\{\tilde{\pi}^s_t\}$ between $n$ submaps, we perform a final optimization to recover the \emph{global} association $\{\pi^s_u\}$, which maps objects in each submap $M^s$ to the \emph{universe} $U$ of objects, i.e., the set of all trees in the forest. 
Recovering $\{\pi^s_u\}$ requires the pairwise associations $\{\tilde{\pi}^s_t\}$ to be \emph{cycle consistent}, i.e., the composition of pairwise mappings along any cycle of submaps should result in the identity mapping \citep{NguyenCGF2011}.

To better understand the importance of cycle consistency, we look at the \emph{data association graph} $\Gcal$ \citep{LeonardosICRA2017} induced by $\{\tilde{\pi}^s_t\}$.\footnote{$\Gcal$ should be distinguished from the correspondence graph defined in the previous section for pairwise association.}
With a slight abuse of notation, we let
each vertex in $\Gcal$ represent an object $o^s_i$ in a submap $M^s$.
Two vertices $o^s_i$, $o^t_j$ are connected by an edge if they are 
matched in the input associations,
i.e., if $\pitilde^s_t(i) = j$.
We note that these edges are inherently \emph{transitive}:
if $o^s_i$ is matched to $o^t_j$
and $o^t_j$ is matched to $o^\tau_k$, 
then it must be true that
$o^s_i$ is matched to $o^\tau_k$.
\cite{TronICCV2017} show that 
$\{\tilde{\pi}^s_t\}$ is cycle consistent 
if and only if the corresponding data association graph consists of disjoint cliques,
and furthermore no two objects from the same submap appear in the same clique. 
In this case,
map fusion happens naturally
by assigning each clique in $\Gcal$
to a unique object in the universe $U$.
Figure~\ref{fig:consistent_induced_graph} provides an example.

In practice, however, $\{\tilde{\pi}^s_t\}$ often contain noisy matches that
violate the cycle consistency principle; an example is shown in Figure~\ref{fig:inconsistent_induced_graph}.
After invoking the transitive property of all edges,
a chain of spurious associations (red edges) 
would incorrectly fuse two objects belonging to the same submap,
which contradicts the
assumption that each submap
contains distinct objects.
Therefore, for map fusion, we must find a cycle consistent set of pairwise associations $\{\pi^s_t\}$ that resembles the input associations $\{\tilde{\pi}^s_t\}$ as much as possible. Given $\{\pi^s_t\}$, it is again straightforward to recover the underlying
global associations $\{\pi^s_u\}$, by 
assigning each clique in the induced data association graph of 
$\{\pi^s_t\}$ to a unique object.
In the computer vision literature,
this procedure is known as
\emph{cycle consistent multiway matching}
\citep{PachauriNIPS2013}.

Among all algorithms proposed for cycle consistent multiway matching, 
we choose CLEAR (Consistent Lifting, Embedding, and Alignment Rectification)
proposed by \cite{FathianCLEAR2019} for its high precision and 
superior speed. 
CLEAR takes as input the aggregate matrix,
\begin{equation}
A = \begin{bmatrix}
0 & \Pitilde_2^1 & \cdots & \Pitilde_n^1 \\
\Pitilde_1^2 & 0 & \cdots & \Pitilde_n^2 \\
\vdots& \vdots& \ddots & \vdots \\
\Pitilde_1^n & \Pitilde_2^n & \cdots & 0 
\end{bmatrix},
\end{equation}
which corresponds to the adjacency matrix of $\Gcal$.
Let $D$ denote the (diagonal) degree matrix of $\Gcal$.
The Laplacian and normalized Laplacian of $\Gcal$ are defined as follows,
\begin{equation}
    L \triangleq D - A, \; L_\text{nrm} \triangleq D^{-\frac{1}{2}} L  D^{-\frac{1}{2}}.
\end{equation}
From here, the algorithm proceeds in two stages. 
In the first stage, the number of objects $m$ in the universe is estimated
from the spectrum of the normalized Laplacian $L_\text{nrm}$.
In the second stage, an eigendecomposition is performed over $L_\text{nrm}$
to obtain an embedding vector $v^s_i \in \mathbb{R}^m$ for each 
vertex $o^s_i$ in $\Gcal$.
These embeddings are then used to cluster
vertices into $m$ disjoint cliques using the Hungarian algorithm. 
\Yedit{}
{
When computing the embedding vectors, we use a block singular value decomposition (SVD) method that leverages the seperable structure of the underlying data association graph.
To further accelerate computation, we also implemented the greedy method for clustering as suggested in \citep{FathianCLEAR2019}.
}

The final global data association $\{\pi^s_u\}$ is obtained 
by assigning vertices in each clique to a unique object in the universe,
i.e., 
$\pi^s_u(i) = c$ if vertex $o^s_i$ belongs to clique $c$.
In addition, $\pi^s_u$ can also be used to recover a cycle consistent pairwise matching
simply via, 
\begin{equation}
    \pi^s_t = \pi^s_u \circ \pi^u_t,
\end{equation}
where $\circ$ denotes the composition operator for permutations.

\subsection{Global Landmark SLAM}
\label{sec:isam}
After solving for the global data association $\{\pi^s_u\}$, we perform SLAM
to refine the estimates of all poses and objects in the fused map.
To do this,
we jointly optimize over the origin $x_t \in \text{SE}(2)$
of each submap $M^t$, 
which also corresponds to the vehicle pose when $M^t$ is initialized, as well as
the positions of the $m$ objects $l_1, \hdots, l_m \in \mathbb{R}^2$ in the environment. 
Recall that $m$ is estimated by CLEAR; see Section~\ref{sec:clear}.

Following the standard formulation of landmark SLAM, 
we consider two types of measurements during optimization. 
First, we use \emph{odometry} measurements $z^{t}_{t+1} \in \text{SE}(2)$
extracted from EKF, which links consecutive submaps belonging to the same vehicle.
For two submaps from different vehicles, an odometry measurement is not available as these vehicles do not share a common coordinate system.
Second, we also consider \emph{observation} measurements
which links a map origin to landmarks observed in this submap.
These measurements can be extracted from the global data association $\{\pi^s_u\}$:
for each object $o^s_i$ in map $M^s$, 
we retrieve the index $c$
of the corresponding tree via
$c = \pi^s_u(i)$. 
An observation is then created between
$x_s$ and $l_c$, with a relative transformation constraint specified by $p^s_i \in \mathbb{R}^2$, where $p^s_i$ is the observed position of this tree in the coordinate frame of map $M^s$.

After all variables and measurements are initialized, 
we use iSAM \citep{KaessiSAM}
to carry out the nonlinear optimization. 
Finally, the optimized map origins $\{x_t\}$
are transmitted back to the vehicles
to correct estimation drifts during local estimation,
by realigning all onboard submaps using the updated estimates of $\{x_t\}$.

\section{Simulation Experiments}
\label{sec:sim_experiments}
\Yedit{
In order to evaluate the performance of the proposed exploration and SLAM algorithms over many trials, we leveraged a high fidelity simulation environment, and used largely the same set of parameters as in the following real-world experiments.}
{In order to evaluate the performance of the proposed tree detection, exploration, and SLAM algorithms over many trials, we 
conducted experiments in simulations and used largely the same set of parameters as in the following real-world experiments.}
Certain parameters were different to compensate for the differences in the simulation and the real world, e.g., the limit on maximum acceleration was set much higher in real-world experiments to compensate for unmodeled external forces such as wind.

\subsection{Simulation Setup}
\Yedit{}{To evaluate tree detection under different settings of sensor noise and forest density (Section~\ref{sec:tree_detection_results}), 
we generated random forests in 2D where tree positions are sampled from a 2D Poisson point process \citep{Chiu2013stochastic} and tree radii are sampled uniformly from $0.1-0.3$ m. 
We then simulated laser range measurements corrupted by zero-mean Gaussian noise for a robot with a $30$~m sensing range, a $270^\circ$ field of view, and a $0.25^\circ$ angular resolution (same as real-world experiments). 
}

\Yedit{}
{To evaluate the proposed exploration and SLAM algorithms (Section~\ref{sec:exploration_results}-\ref{sec:slam_results}), 
we also leveraged a high fidelity simulation environment based on the Unity game engine.}
Vehicle dynamics and IMU readings for a simulated vehicle were generated using the Drake toolbox \citep{tedrake2014drake} and a quadrotor model as described in \citep{mellinger2012trajectory}. To test integration with the full stack, we also utilized the Pixhawk \citep{meier2012pixhawk} Software-In-The-Loop (SITL) to simulate the motor commands from the Pixhawk, which were fed back into the Drake dynamics model. We used the Unity game engine to simulate 2D laser scans in a randomly generated forest environment at roughly $30$ Hz with a $270^\circ$ field of view and 30 m range. All sensor measurements were simulated with low noise.
The simulated setpoints generated by the motion planner were passed to the Pixhawk SITL. 

\Yedit{}
{
\subsection{Tree Detection Results in Simulation}
\label{sec:tree_detection_results}

\begin{figure*}[h]
	\centering
	\begin{subfigure}{0.32\textwidth}
	\includegraphics[width=\textwidth]{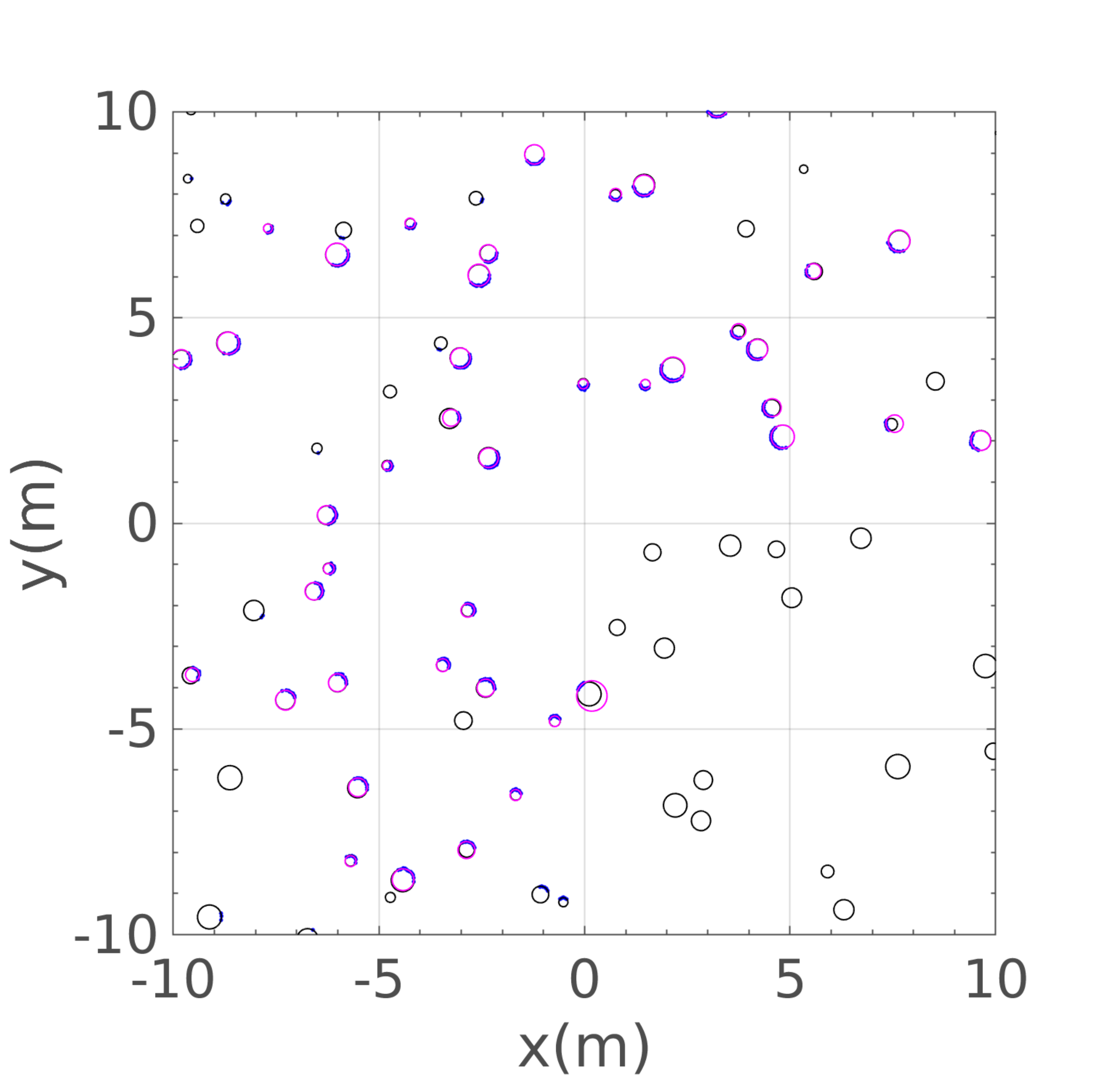}
	\caption{Example random forest in 2D.}
	\label{fig:simulated_scan}
	\end{subfigure}
	\begin{subfigure}{0.32\textwidth}
	\includegraphics[width=\textwidth]{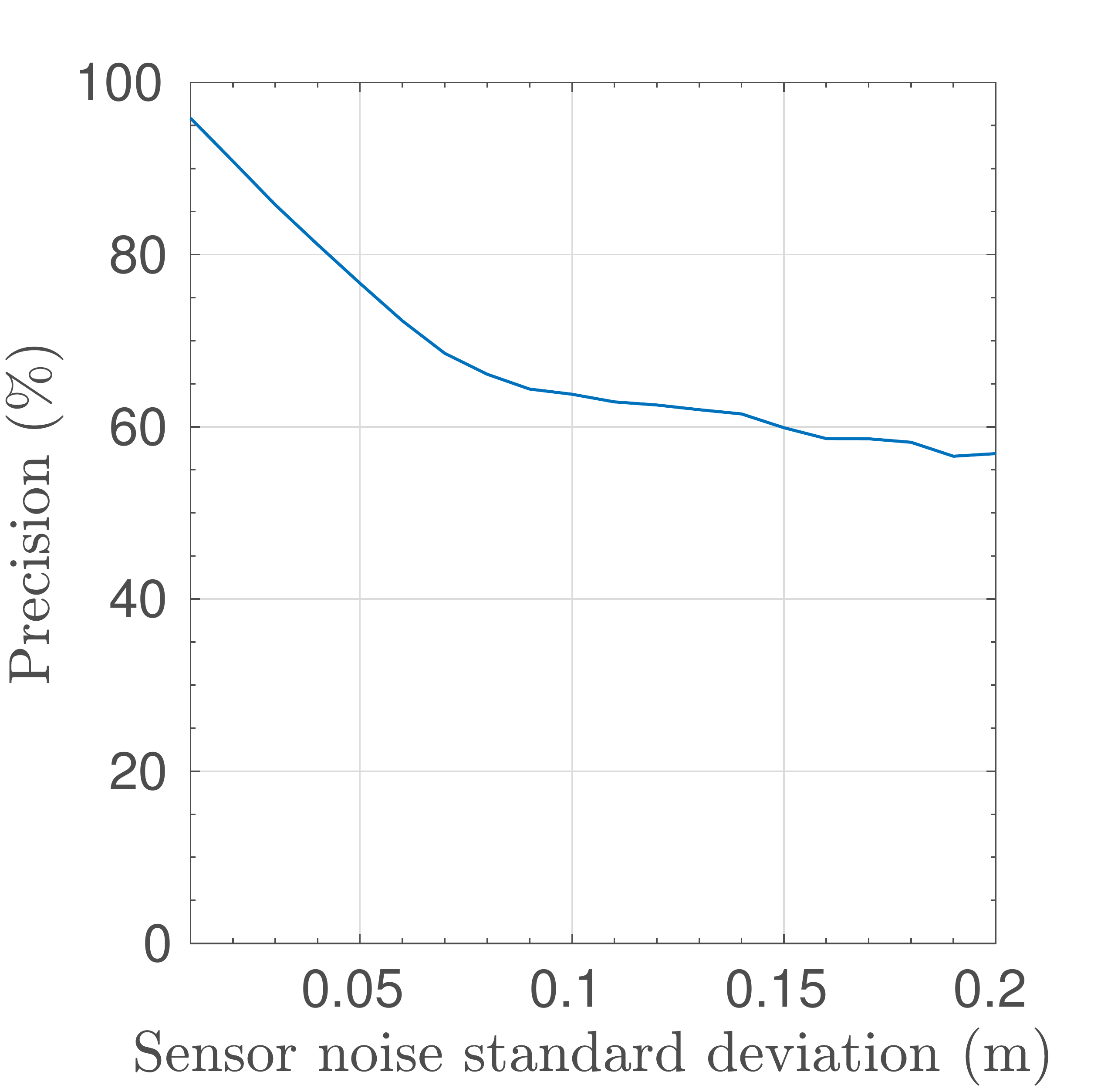}
	\caption{Precision vs. sensor noise}
	\label{fig:tree_detection_vary_noise}
    \end{subfigure}
	\begin{subfigure}{0.32\textwidth}
    \includegraphics[width=\textwidth]{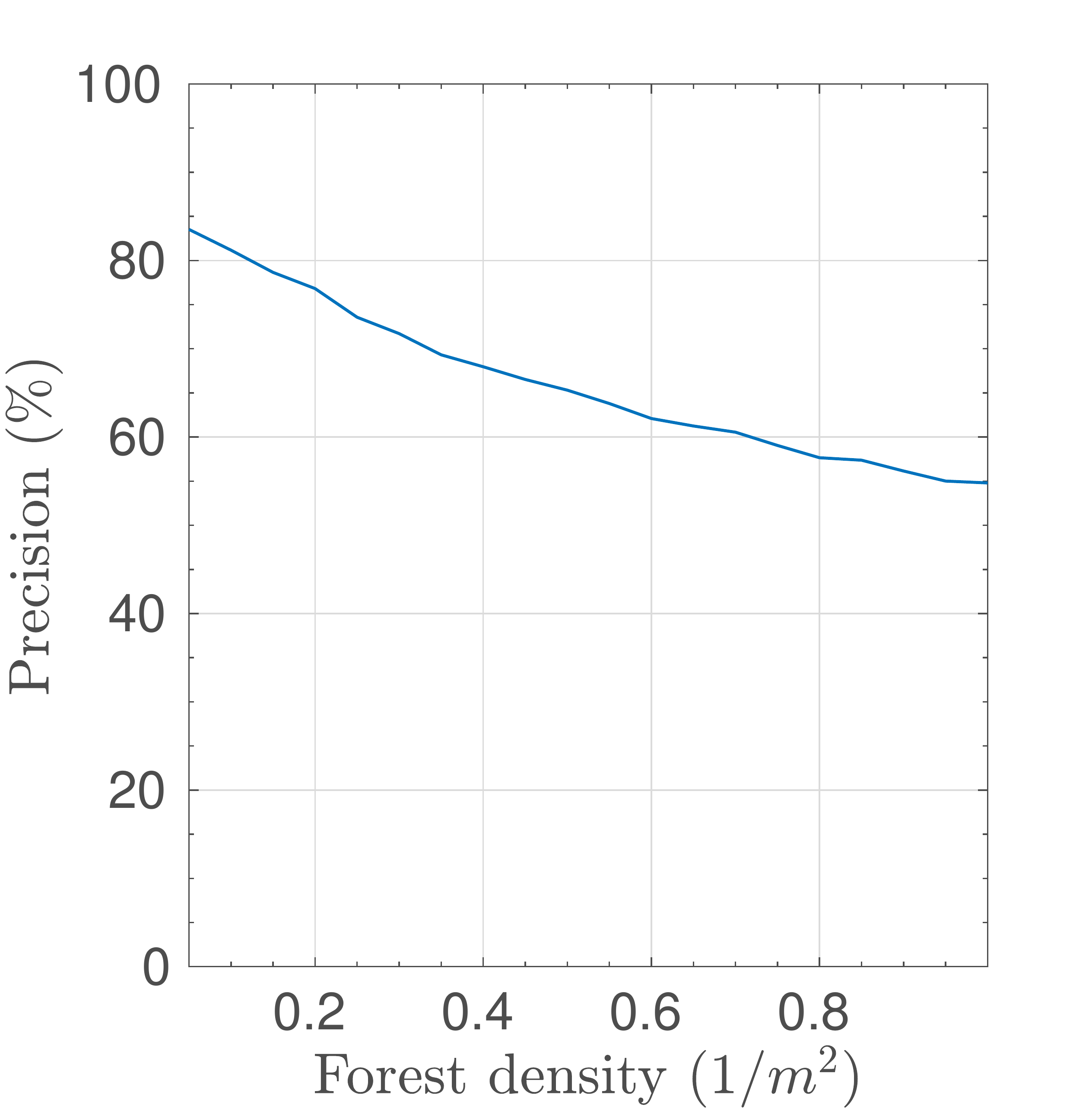}
	\caption{Precision vs. forest density}
	\label{fig:tree_detection_vary_density}
	\end{subfigure}
	\caption{ \small
	\Yedit{}{
	(a) Example simulation of a 2D random forest. 
	The robot is placed at the origin, with a $270^\circ$ field of view that does not cover the fourth quadrant.  
	Black circles denote ground truth tree trunks. 
	Blue dots show simulated laser returns.
	Magenta circles show detected tree trunks.
	(b) Precision of tree detection with increasing sensor noise, under fixed forest density of $0.2/\text{m}^2$. 
	(c) Precision of tree detection under increasing forest density, with sensor noise standard deviation fixed at $0.05$~m.
	For each level of noise and density, we performed 500 random simulations and recorded the average precision.}}
	\label{fig:tree_detection_results}
\end{figure*}

Figure~\ref{fig:simulated_scan} shows an example random forest in 2D together with the tree trunks detected by our algorithm. 
Using the simulator, we evaluated the precision of tree detection with respect to increasing sensor noise and forest density (i.e., average number of trees per unit area).
For each level of noise and density, 
we performed 500 random simulations and recorded the average precision.
A tree detection is classified as a true positive 
if the estimated position is within $0.5$~m of ground truth and
the error of estimated radius is less than $30\%$.
As shown in Figure~\ref{fig:tree_detection_vary_noise},
the precision of tree detection tends to be more sensitive under low noise and less so under higher noise. 
On the other hand, 
precision also degrades as the forest density increases, as shown in Figure~\ref{fig:tree_detection_vary_density}.
We observed that many failure cases in dense forests are caused by 
the Dirichlet process clustering algorithm incorrectly merging points belonging to nearby trees. 
In these cases, more sophisticated tree detection methods are required to further improve robustness. 
}

\Yedit{}
{
\subsection{Loop Closure Candidates Detection Results in Simulation}
Using the same simulator as previous section, 
we evaluated the effectiveness of GLARE and GLAROT (Section~\ref{sec:glare}) at detecting loop closure candidates, under increasing forest density.
We recorded the average GLAROT distances for scan pairs that either share sufficient overlap in the fields of view (and hence likely to be loop closures), or do not share any overlap in the fields of view. 
In Figure~\ref{fig:glare_results}, we refer to the former as ``true pairs'' and the latter as  ``false pairs''. 
Intuitively, the larger the gap is between the GLAROT distances of true pairs and false pairs, 
the more effective GLAROT is at identifying loop closures candidates. 
In our experiments, we used the default discretization of $N_\rho = 120, N_\theta = 12$, and a reduced Gaussian blur with standard deviation $0.01$.
We observed that moderate change in discretization does not lead to significant change in performance. 

As expected, 
in Figure~\ref{fig:glare_results}, the GLAROT distances for true pairs (i.e., loop closures candidates) is consistently lower than false pairs, with a reasonable margin under all levels of forest density. 
In particular, the margin between the two curves is larger in sparser forests, indicating that GLAROT is more effective in this case. 
This is because as the forest gets denser, occlusions happen more frequently, effectively reducing the sensor's field of view. 
We note that this issue can be mitigated by aggregating information from multiple scans, which is already implemented in our current system via the use of submaps.

\begin{figure}[h]
	\centering
	\includegraphics[width=0.45\textwidth]{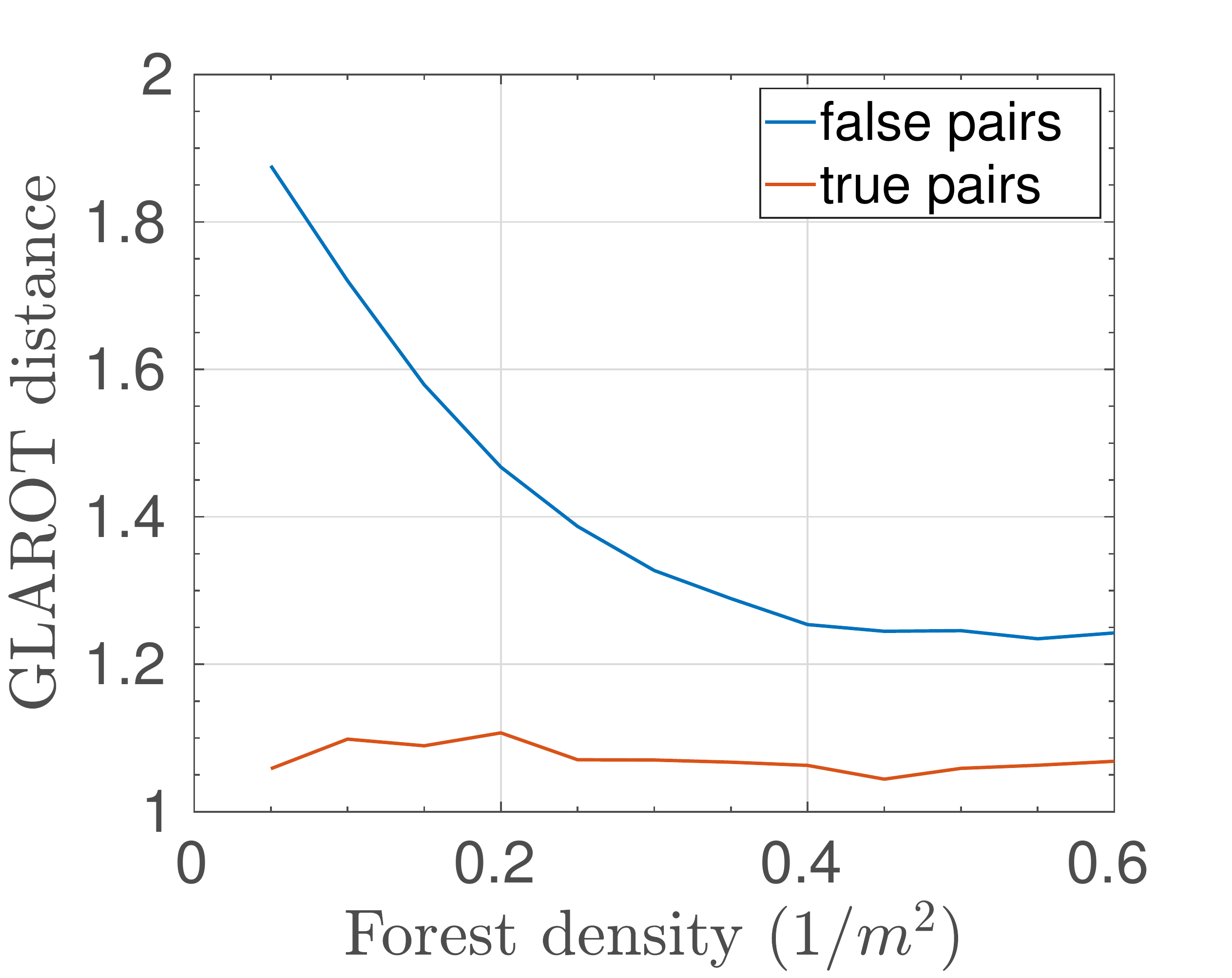}
	\caption{ \small
	\Yedit{}{
	Average GLAROT distances for simulated scan pairs that either have sufficient overlap in the fields of view (true pair) or not (false pair), as a function of increasing forest density.
	Each data point is the result of averaging over 100 random simulations.
	The margin between the GLAROT distances is larger in sparser forests, indicating that GLAROT is more effective at identifying loop closure candidates in this case.
	We note that results under density level larger than $0.6 / m^2$ is omitted, as the margin between the two curves stays relatively constant.}}
	\label{fig:glare_results}
\end{figure} 

}

\subsection{Exploration Results in Simulation}
\label{sec:exploration_results}
\begin{figure*}[h]
    \vspace{15pt}
	\centering
    \begin{subfigure}{0.24\textwidth}
	\includegraphics[width=1.0\textwidth]{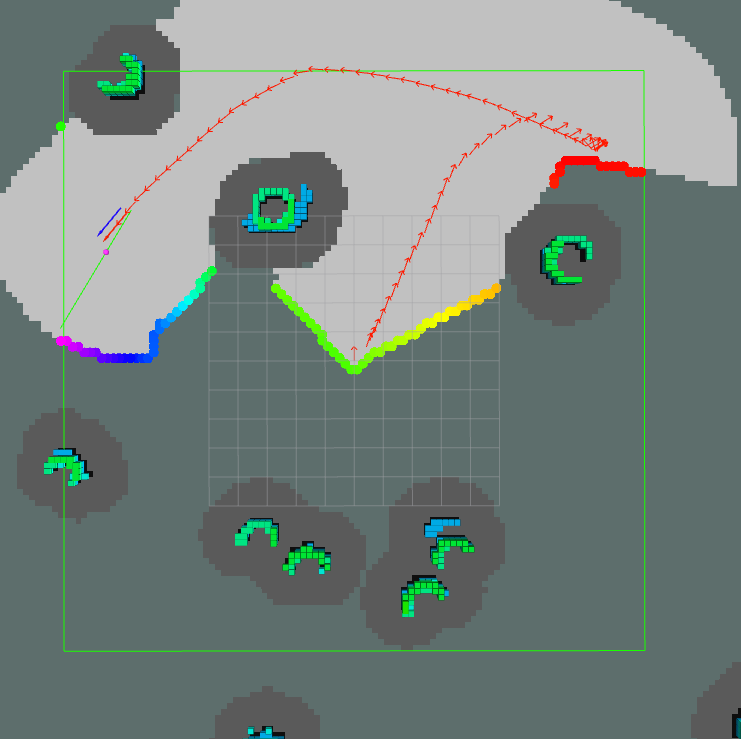}
	\caption{$t=70$ sec\\\hspace{1in}}
	\label{fig:sim_proposed_70}
    \end{subfigure}
    \begin{subfigure}{0.24\textwidth}
	\includegraphics[width=1.0\textwidth]{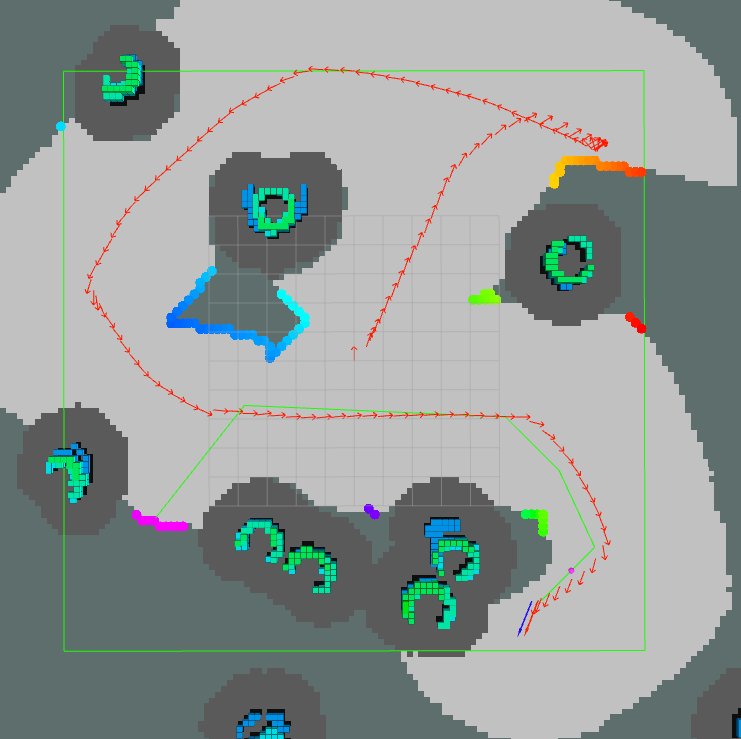}
	\caption{$t=130$ sec\\\hspace{1in}}
    \end{subfigure}
	\begin{subfigure}{0.24\textwidth}
	\includegraphics[width=1.0\textwidth]{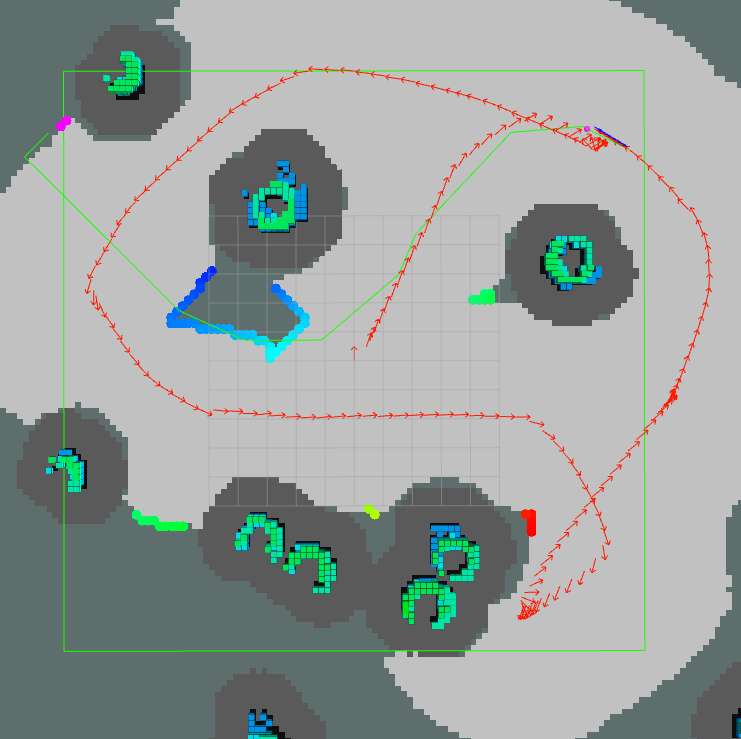}
	\caption{$t=180$ sec\\\hspace{1in}}
	\end{subfigure}
	\begin{subfigure}{0.24\textwidth}
	\includegraphics[width=1.0\textwidth]{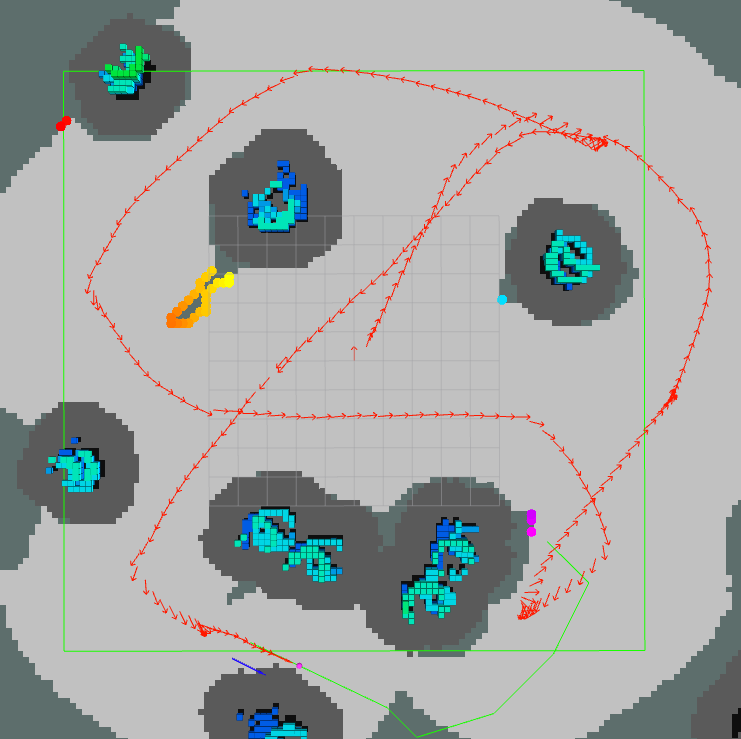}
	\caption{$t=250$ sec\\\hspace{1in}}
	\label{fig:sim_proposed_250}
	\end{subfigure}\\
	\begin{subfigure}{0.24\textwidth}
	\includegraphics[width=1.0\textwidth]{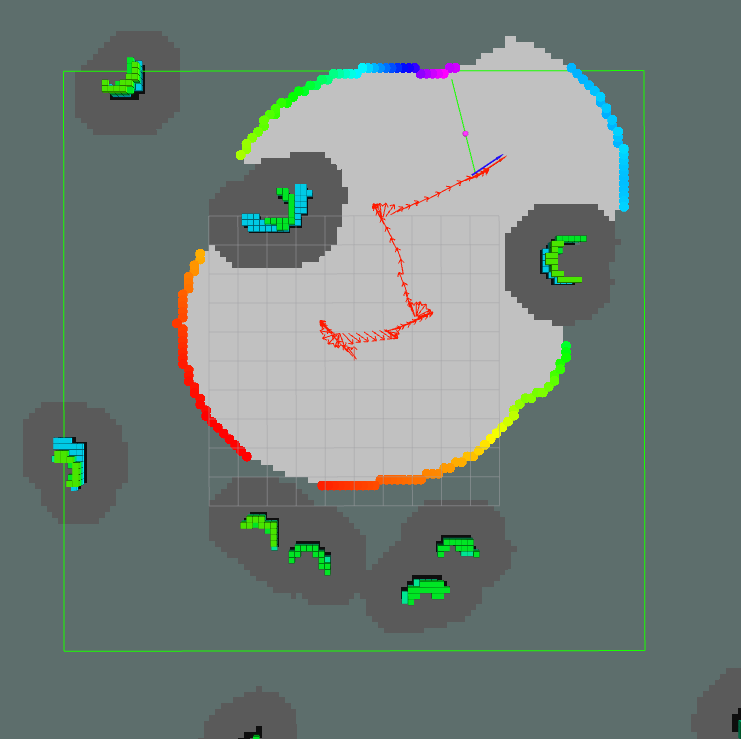}
	\caption{$t=70$ sec\\\hspace{1in}}
	\label{fig:sim_baseline_70}
    \end{subfigure}
    \begin{subfigure}{0.24\textwidth}
	\includegraphics[width=1.0\textwidth]{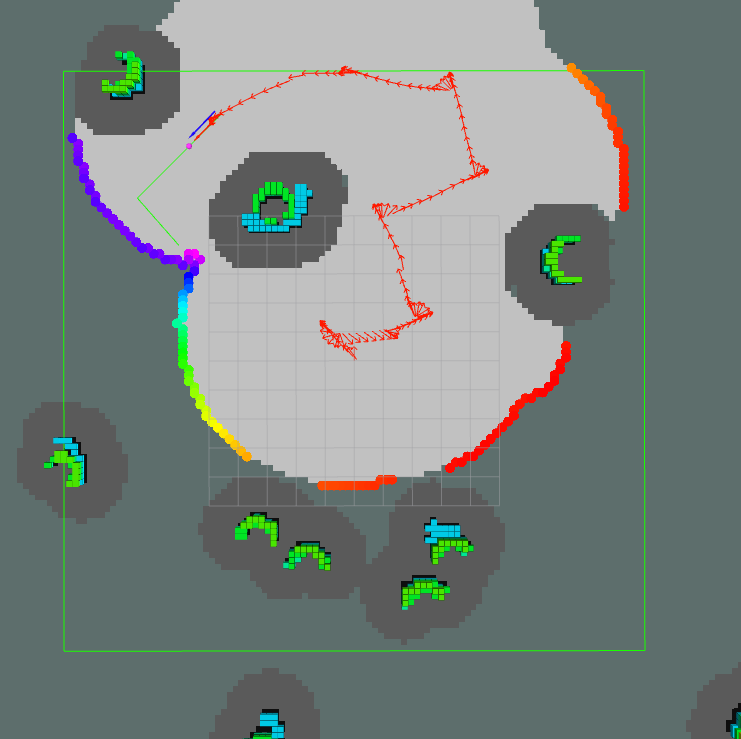}
	\caption{$t=130$ sec\\\hspace{1in}}
    \end{subfigure}
	\begin{subfigure}{0.24\textwidth}
	\includegraphics[width=1.0\textwidth]{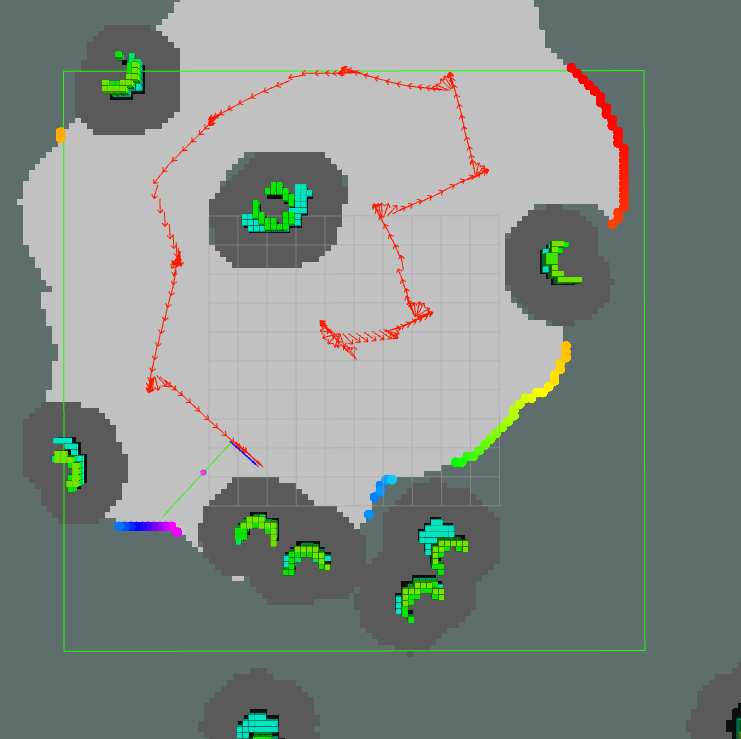}
	\caption{$t=180$ sec\\\hspace{1in}}
	\end{subfigure}
	\begin{subfigure}{0.24\textwidth}
	\includegraphics[width=1.0\textwidth]{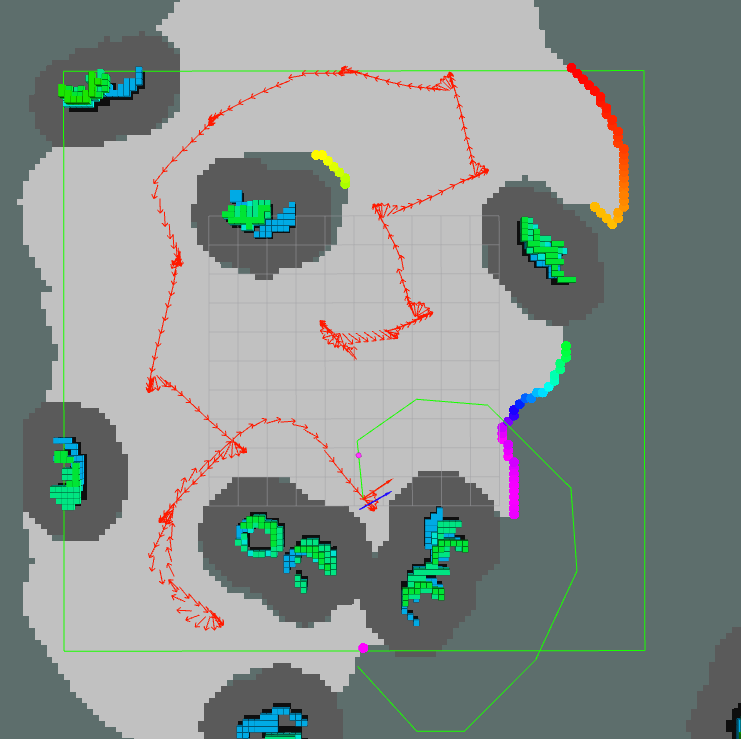}
	\caption{$t=250$ sec\\\hspace{1in}}
	\label{fig:sim_baseline_250}
	\end{subfigure}
	\caption{Vehicle trajectories and partial maps in simulation at different time steps (seconds). \textbf{(a)}-\textbf{(d)} show the proposed planner; \textbf{(e)}-\textbf{(h)} show the baseline planner. Green rectangle denotes the search area assigned to the vehicle. Colored point clouds shows the set of all frontiers with colors representing different costs (magenta to red in increasing cost). 
	Red path shows the trajectory taken by the vehicle. 
    The trajectory in \textbf{(a)}-\textbf{(d)} is smoother and more time-efficient compared to the trajectory in \textbf{(e)}-\textbf{(h)}.
    At $t=250$ sec, the proposed planner nearly completed the mission, while the search area was only partially explored by the baseline planner.}
	\label{fig:sim_trajectory}
\end{figure*}

We evaluated the proposed frontier-based exploration planner in simulation.
A single UAV was tasked with covering a $20\text{m} \times 20\text{m}$ search area in a randomly generated forest. 
To benchmark our proposed planner, we also implemented 
the classical frontier selection algorithm that greedily selected the closest frontier in terms of the Euclidean distance \citep{YamauchiICRA1998}.
For comprehensive evaluation, different search areas with varying search difficulty (e.g., density of obstacles) are assigned to the vehicle. 
For each search area, multiple exploration missions were carried out and average performances are recorded in Table~\ref{table:sim_results}. 
\Yedit{}{
Since our system is developed for search and rescue scenarios, 
the most critical performance metric is the time it takes for the vehicle to complete the exploration mission.
As shown in Table~\ref{table:sim_results}, }
the proposed planner clearly outperformed the baseline planner in terms of both the total time to complete the mission and the average speed during flight. 
Figure~\ref{fig:sim_trajectory}
shows the trajectory of the proposed planner \ref{fig:sim_proposed_70}-\ref{fig:sim_proposed_250} and the baseline planner \ref{fig:sim_baseline_70}-\ref{fig:sim_baseline_250} in an example search area. 
As expected, the proposed planner produced a much smoother overall trajectory compared to the baseline and completed the mission in a shorter period of time.

\begin{table}[tb]
\caption{Comparison with baseline planner in three different search areas in simulation. Each planner was evaluated multiple times inside each area and average completion time and flight speed were recorded. In all three cases, the proposed planner was able to maintain a higher velocity while covering the search area more efficiently, resulting in a much shorter average exploration time compared to the baseline planner.}
\vspace{.1in}
\label{table:sim_results}
\centering
\begin{tabular}{|c|c|c|c|c|}
\hline
Area & Planner & Duration (sec) & Avg. Speed (m/sec) \\ \hline 
1 & Proposed & $\mathbf{313.50}$         &  $\mathbf{0.84}$         \\ \hline
1 & Baseline & $504.87$         & $0.69$         \\ \hline
2 & Proposed & $\mathbf{340.52}$         & $\mathbf{0.82}$         \\ \hline
2 & Baseline & $448.91$         & $0.71$         \\ \hline
3 & Proposed & $\mathbf{302.41}$         & $\mathbf{0.80}$         \\ \hline
3 & Baseline & $477.96$         & $0.72$         \\ \hline
\end{tabular}
\end{table}

\subsection{SLAM Results in Simulation}
\label{sec:slam_results}
We evaluated the capability of our submap-based map representation (Section~\ref{sec:onboard_estimation}) to correct onboard estimation drifts. 
\Yedit{}{In practice, such capability is needed to make sure that the final map produced by the ground station is as accurate as possible. 
Similar to the previous section, }
the UAV was tasked with exploring a $20\text{m} \times 20\text{m}$ search area in simulation.
\Yedit{which was different from the one used in the previous section}
The vehicle initialized a new submap every $5$ seconds. 

\begin{figure*}[h]
\vspace{8pt}
\centering
\begin{subfigure}{0.49\textwidth}
\includegraphics[width=1\textwidth]{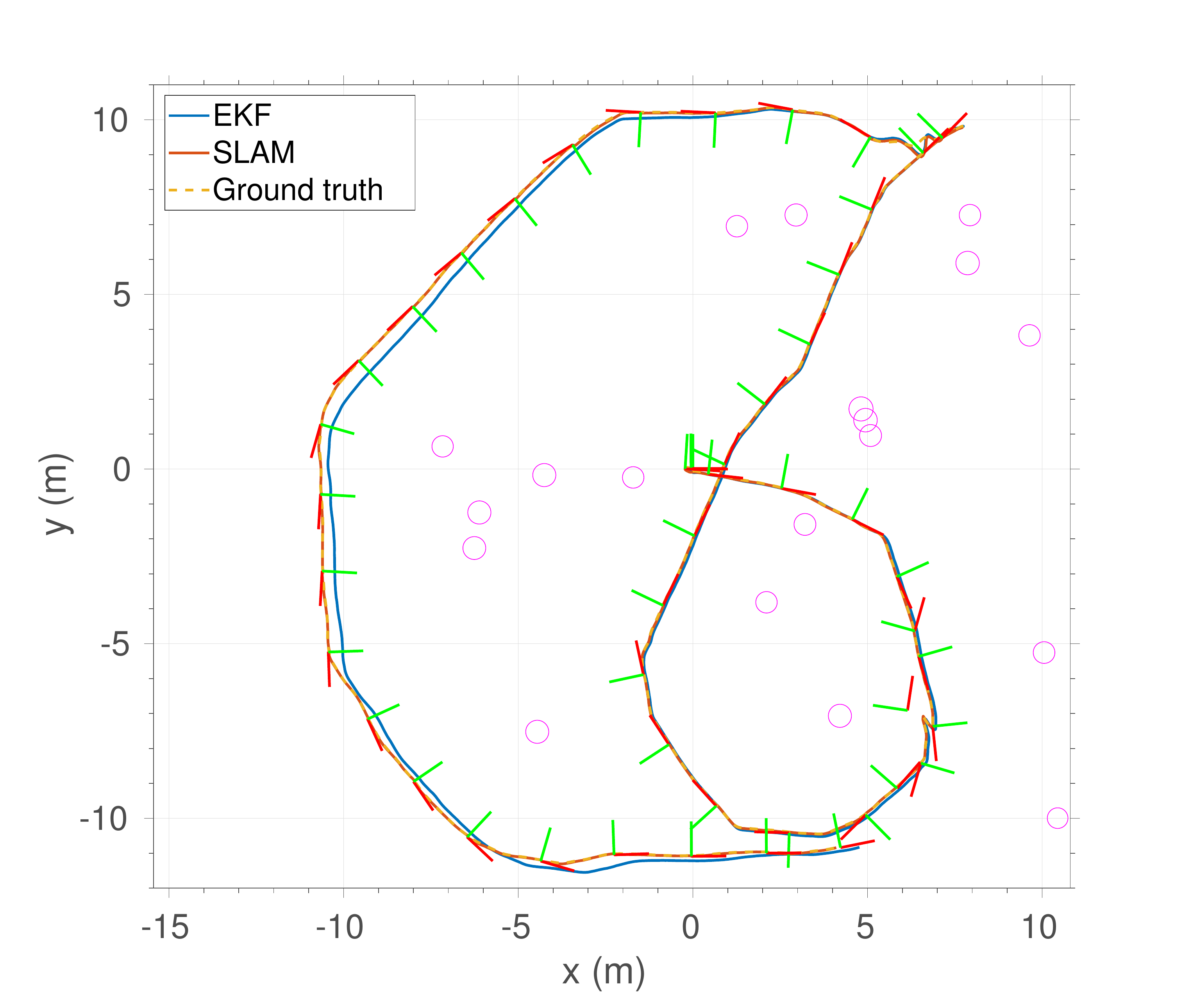}
\caption{Trajectory estimates in simulation}
\label{fig:slam_sim_trajectory}
\end{subfigure} 
\begin{subfigure}{0.49\textwidth}
\includegraphics[width=1\textwidth]{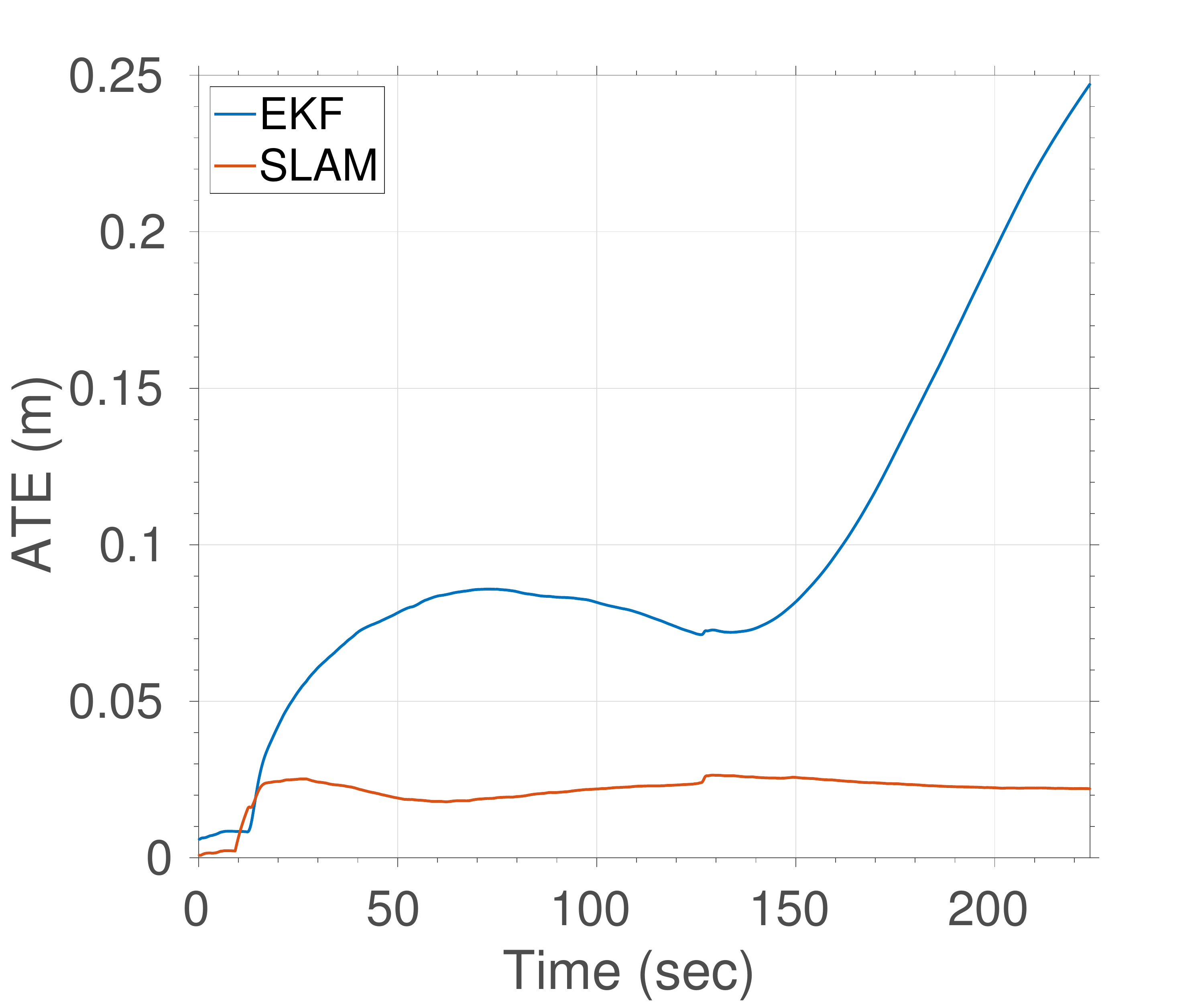}
\caption{Absolute trajectory error (ATE) in simulation}
\label{fig:slam_sim_ate}
\end{subfigure}
\caption{
\textbf{(a)} 
Vehicle trajectory estimated by EKF (blue) and SLAM (red) in simulation. The ground truth trajectory is shown in yellow, which mostly overlaps with the SLAM estimates. Each coordinate frame shows the optimized origin of a submap. Each magenta circle represents a tree optimized during landmark SLAM. 
\textbf{(b)}
Absolute trajectory errors (ATE) \citep{SturmATE}, defined as the average distance between ground truth poses and estimated poses, as a function of exploration time. While the ATE associated to EKF exceeded $0.25$~m, performing landmark SLAM bounded the ATE below $0.03$~m. The results clearly demonstrate the ability of the proposed SLAM pipeline to correct onboard estimation drifts. 
}
\label{fig:slam_sim}
\vspace{25pt}
\end{figure*}

Figure~\ref{fig:slam_sim_trajectory} shows the ground truth trajectory (yellow) and the trajectory estimated by EKF (blue). As expected, the EKF estimates suffered from accumulated estimation drifts. To correct the drifts, we applied our SLAM pipeline described in Section~\ref{sec:offboard_fusion} to optimize the origins of individual submaps, and refined the estimated trajectory based on the optimized submap origins.\footnote{We did not report CSLAM results in simulation, because our simulation currently only supports single vehicle exploration. Nevertheless, when evaluating estimation accuracy, the number of vehicles is nearly irrelevant because we can always assume the trajectories are produced by a single vehicle. Also, see Section~\ref{sec:planning_and_cslam_results} for real-world CSLAM results with two UAVs.} The resulting trajectory (red) mostly matched the ground truth, confirming the accuracy of the SLAM solution. 

To provide additional quantitative results, we also 
\Yedit{recorded the evolution of the absolute trajectory error (ATE) \citep{SturmATE}, defined as the average distance between ground truth poses and estimated poses.}{use the absolute trajectory error (ATE) \citep{SturmATE}, defined as the average distance between ground truth poses and estimated poses, as the performance critical metric to evaluate our system. The results are shown in Figure~\ref{fig:slam_sim_ate}.}
Due to estimation drifts, the ATE associated to EKF eventually exceeded $0.25$~m; in contrast, performing SLAM effectively bounded the ATE below $0.03$~m.

\section{Flight Experiments}
\label{sec:real_experiments}
Real flight tests were performed in the forest at NASA LaRC, shown in Figure~\ref{fig:intro}.
To test the proposed real-time planning and CSLAM algorithms, we deployed a team of two quadrotors to perform the multi-agent mapping and coverage task. We report qualitative planning and CSLAM performance, as well as analysis of data payload size and algorithm runtimes (Section \ref{sec:planning_and_cslam_results}).
We note that the reported real-time planning results were the same as in our previous work \citep{TianISER2018}, with the frontier-based planning occurring in a single onboard map. Nevertheless, all CSLAM results were updated to incorporate the improvements (notably the submap-based representation and cycle consisteny multiway matching) made in this work. 
Additionally, we performed extensive offline analysis on the global data association methods described in this work using the data collected by a single vehicle from the same experiment (Section \ref{sec:data_association_results}).

\subsection{Outdoor Flight Setup}
\label{sec:outdoor_setup}
Each vehicle was a modified DJI F450 carrying
a horizontally mounted Hokuyo UTM-30LX laser rangefinder, a Pixhawk PX4 unit
providing inertial measurements and motor commands, a downward-facing LidarLite for altitude
measurements, and an Intel NUC computer for onboard computation. The Hokuyo produced laser measurements
at a rate of 40 Hz over an angular field of view of 270$^\circ$ with
0.25$^\circ$ angular resolution. The inertial measurements and LidarLite
measurements were processed at 100 Hz. 
The vehicles were commanded to fly at 1.8 m altitude with a maximum velocity of 2.0~m/sec, and a maximum acceleration of 0.4~m/sec$^2$. 
Communication with the ground station was maintained via 5.8 GHz WiFi; alternatives for peer-to-peer (P2P) communication, e.g., Long Range (LoRa) radios, exist for larger search regions. 
A new submap was initialized on each vehicle after $5$ seconds.
Other parameters used during onboard and offboard operations were:
$\lambda = 0.5$ (Section~\ref{sec:frontier}),
$\tau_\text{cull} = 3$ (Section~\ref{sec:compression}),
$\epsilon_\text{GLAROT} = 1.5$ (Section~\ref{sec:glare}), 
$\epsilon_\text{CG} = 0.15$ (Section~\ref{sec:cg}), and $\tau_\text{CG} = 7$ (Section~\ref{sec:cg}).

\subsection{Real-time Planning and CSLAM Results}
\label{sec:planning_and_cslam_results}

\begin{figure*}
	\centering
    \begin{subfigure}{0.24\textwidth}
	\includegraphics[width=1.0\textwidth]{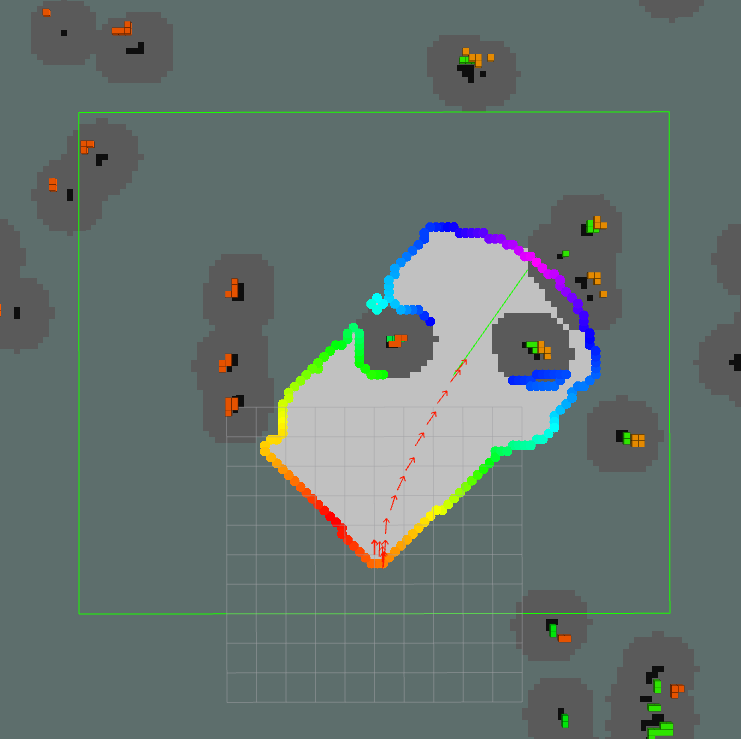}
	\caption{$t = 30$ sec\\\hspace{1in}}
    \end{subfigure}
    \begin{subfigure}{0.24\textwidth}
	\includegraphics[width=1.0\textwidth]{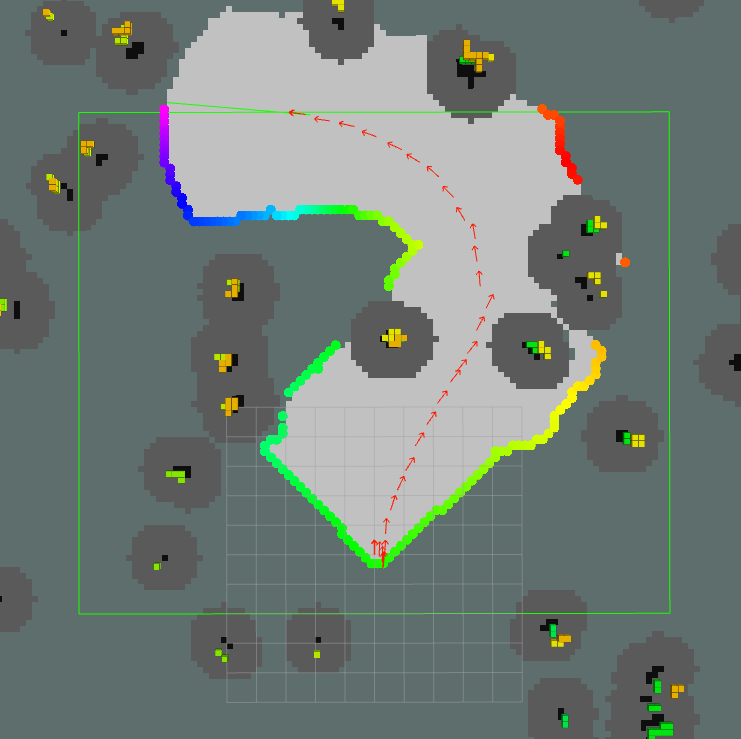}
	\caption{$t = 45$ sec\\\hspace{1in}}
    \end{subfigure}
	\begin{subfigure}{0.24\textwidth}
	\includegraphics[width=1.0\textwidth]{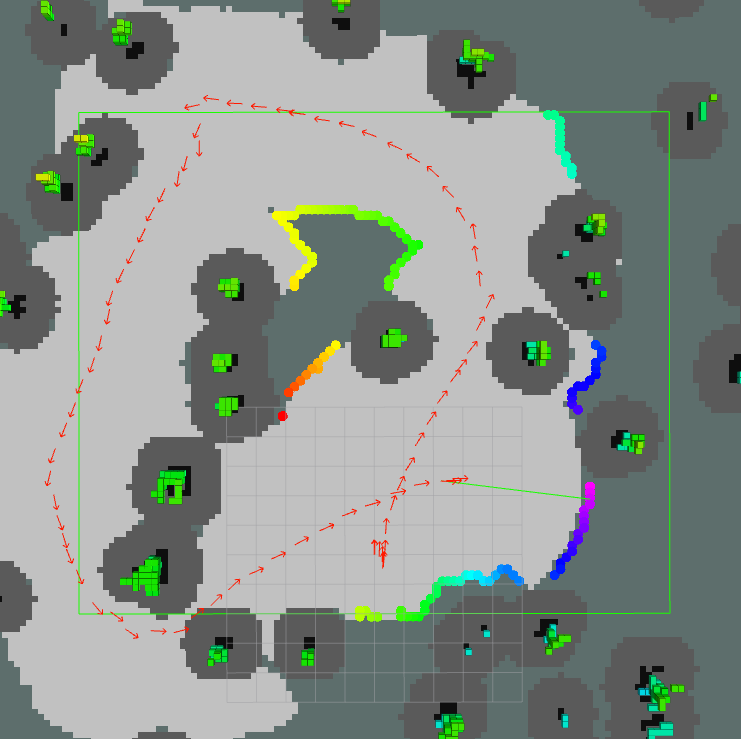}
	\caption{$t = 90$ sec\\\hspace{1in}}
	\end{subfigure}
	\begin{subfigure}{0.24\textwidth}
	\includegraphics[width=1.0\textwidth]{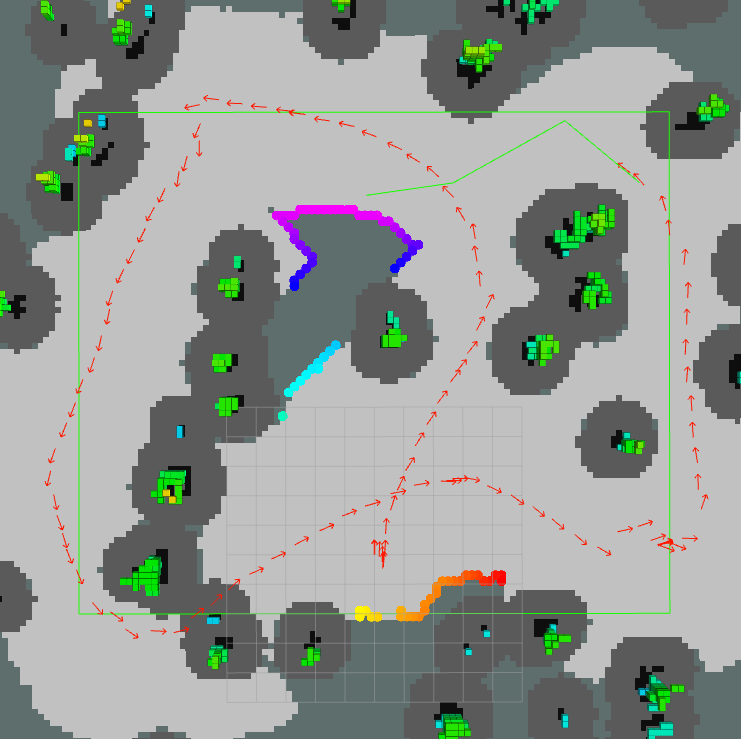}
	\caption{$t = 120$ sec\\\hspace{1in}}
	\end{subfigure}\\
	\begin{subfigure}{0.24\textwidth}
	\includegraphics[width=1.0\textwidth]{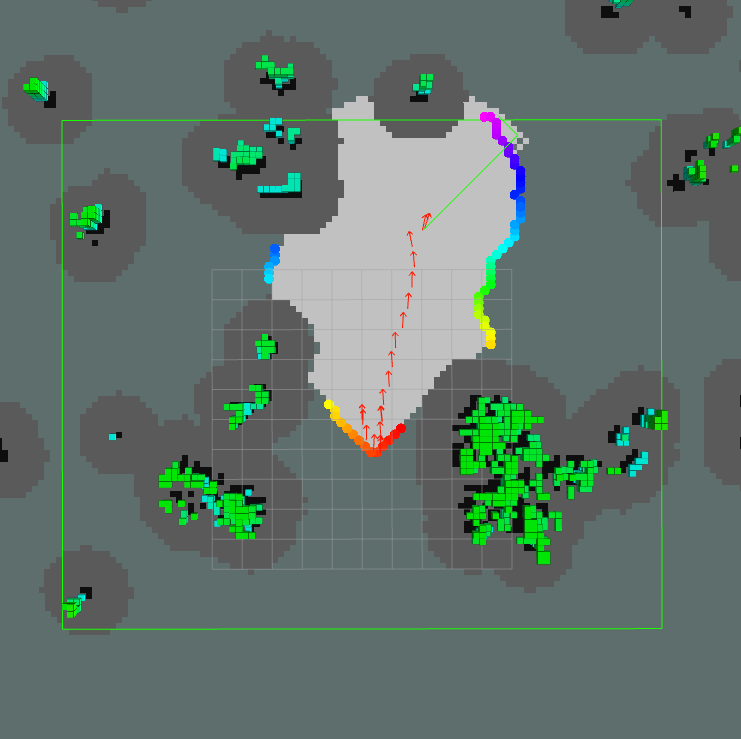}
	\caption{$t = 30$ sec\\\hspace{1in}}
    \end{subfigure}
    \begin{subfigure}{0.24\textwidth}
	\includegraphics[width=1.0\textwidth]{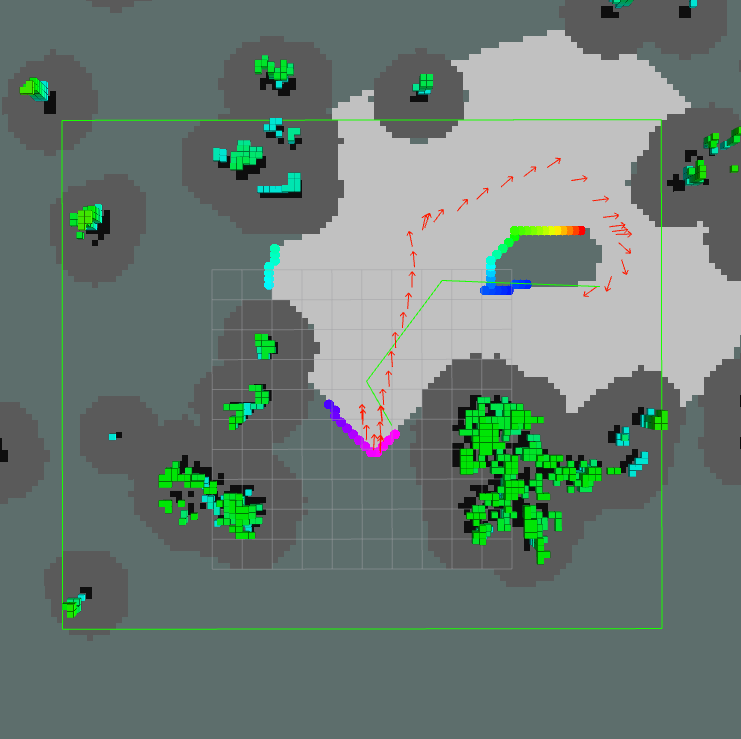}
	\caption{$t = 45$ sec\\\hspace{1in}}
    \end{subfigure}
	\begin{subfigure}{0.24\textwidth}
	\includegraphics[width=1.0\textwidth]{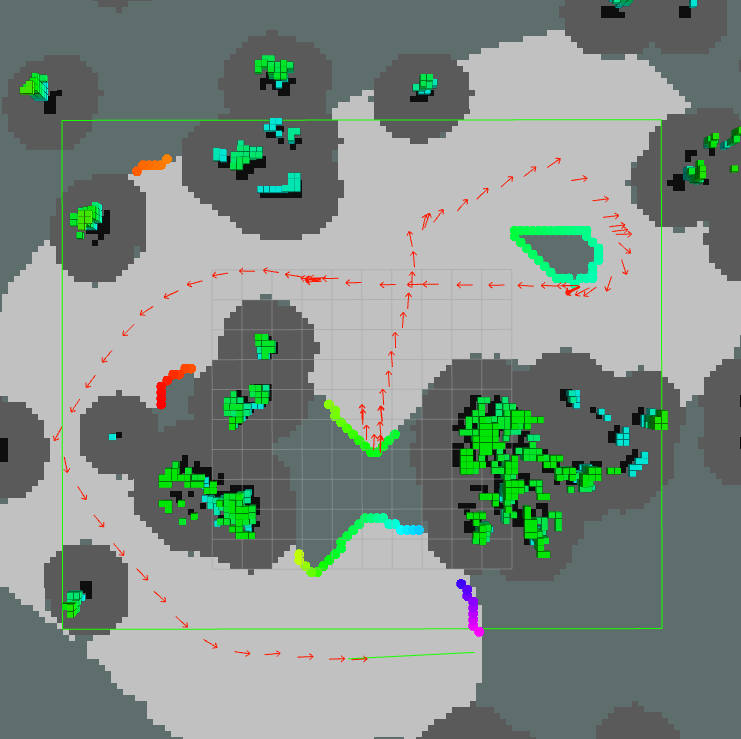}
	\caption{$t = 90$ sec\\\hspace{1in}}
	\end{subfigure}
	\begin{subfigure}{0.24\textwidth}
	\includegraphics[width=1.0\textwidth]{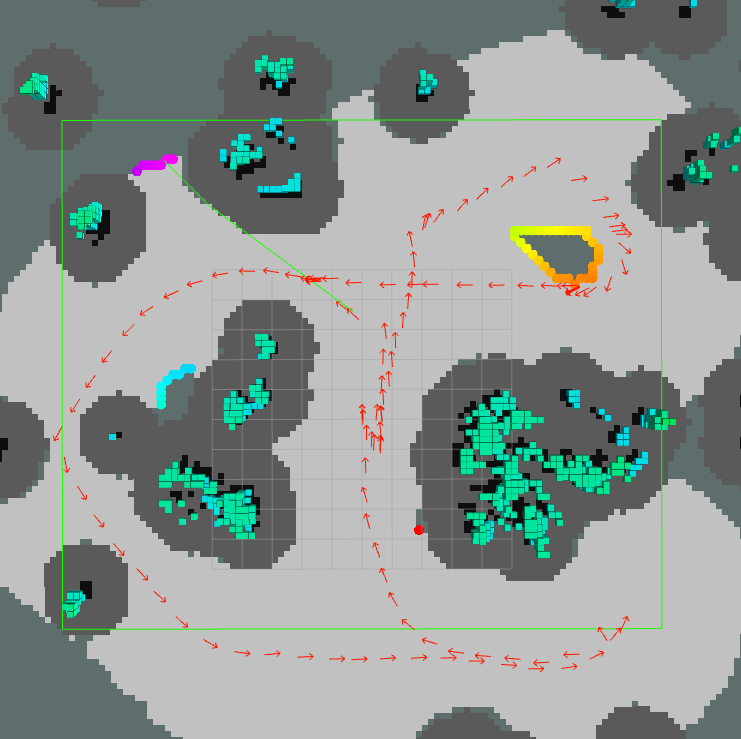}
	\caption{$t = 120$ sec\\\hspace{1in}}
	\end{subfigure}
	\caption{Vehicle trajectories and partial maps in real flight tests at different time steps (seconds). \textbf{(a)}-\textbf{(d)} show the trajectory of the first vehicle and \textbf{(e)}-\textbf{(h)} show the second vehicle. Since the proposed planner preferred frontiers with smaller orientation change, the flight trajectory exhibited bias towards a natural spiral pattern. }
	\label{fig:real_trajectory}
\end{figure*} 

\begin{figure*}
	\centering
    \begin{subfigure}{0.49\textwidth}
	\includegraphics[width=1.0\textwidth]{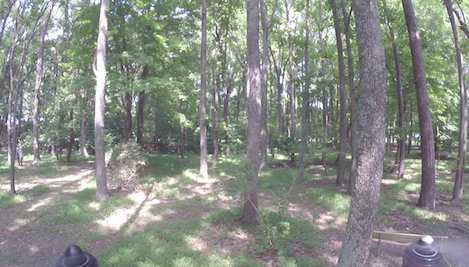}
	\caption{Vehicle 1 at $t = 30$ sec\\\hspace{1in}}
	\end{subfigure}
	\begin{subfigure}{0.49\textwidth}
	\includegraphics[width=1.0\textwidth]{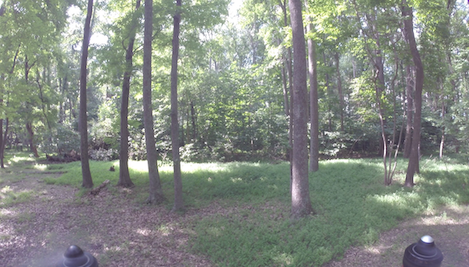}
	\caption{Vehicle 1 at $t = 45$ sec\\\hspace{1in}}
    \end{subfigure}
	\begin{subfigure}{0.49\textwidth}
	\includegraphics[width=1.0\textwidth]{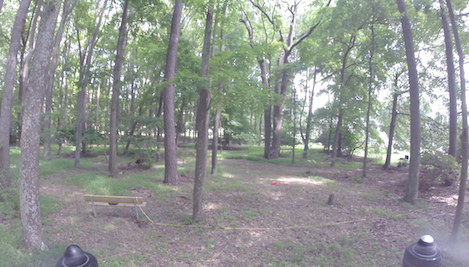}
	\caption{Vehicle 1 at $t = 90$ sec\\\hspace{1in}}
	\end{subfigure}
	\begin{subfigure}{0.49\textwidth}
	\includegraphics[width=1.0\textwidth]{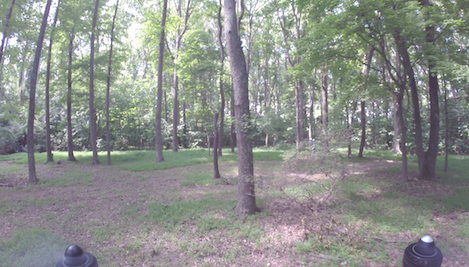}
	\caption{Vehicle 1 at $t = 120$ sec\\\hspace{1in}}
	\end{subfigure}	
    \begin{subfigure}{0.49\textwidth}
	\includegraphics[width=1.0\textwidth]{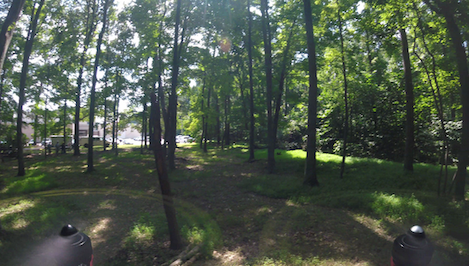}
	\caption{Vehicle 2 at $t = 30$ sec\\\hspace{1in}}
    \end{subfigure}
    \begin{subfigure}{0.49\textwidth}
	\includegraphics[width=1.0\textwidth]{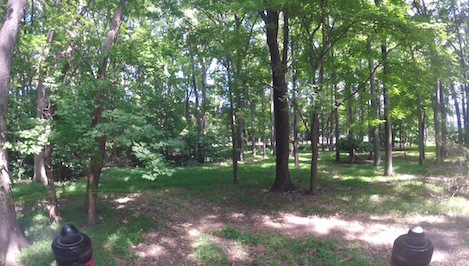}
	\caption{Vehicle 2 at $t = 45$ sec\\\hspace{1in}}
    \end{subfigure}
	\begin{subfigure}{0.49\textwidth}
	\includegraphics[width=1.0\textwidth]{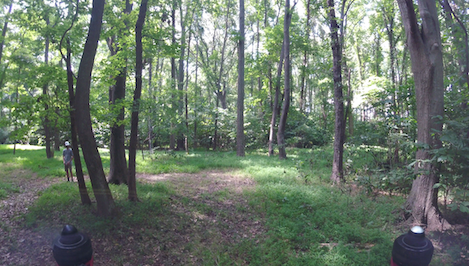}
	\caption{Vehicle 2 at $t = 90$ sec\\\hspace{1in}}
	\end{subfigure}
	\begin{subfigure}{0.49\textwidth}
	\includegraphics[width=1.0\textwidth]{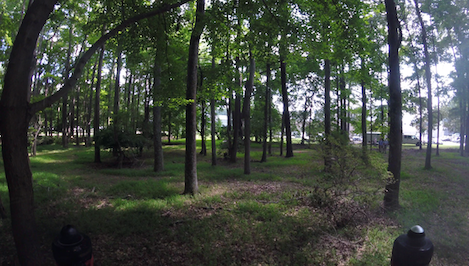}
	\caption{Vehicle 2 at $t = 120$ sec\\\hspace{1in}}
	\end{subfigure}
	\caption{Images from onboard GoPro camera at different time steps (seconds); \textbf{(a)}-\textbf{(d)} show images from the first vehicle and \textbf{(e)}-\textbf{(h)} show images from the second vehicle. Frames are approximately aligned with the vehicle trajectories shown in Figure~\ref{fig:real_trajectory}. } \vspace{0.5cm}
	\label{fig:real_trajectory_gopro}
\end{figure*} 

\begin{figure*}
	\centering
    \begin{subfigure}{0.42\textwidth}
	\includegraphics[width=1.0\textwidth]{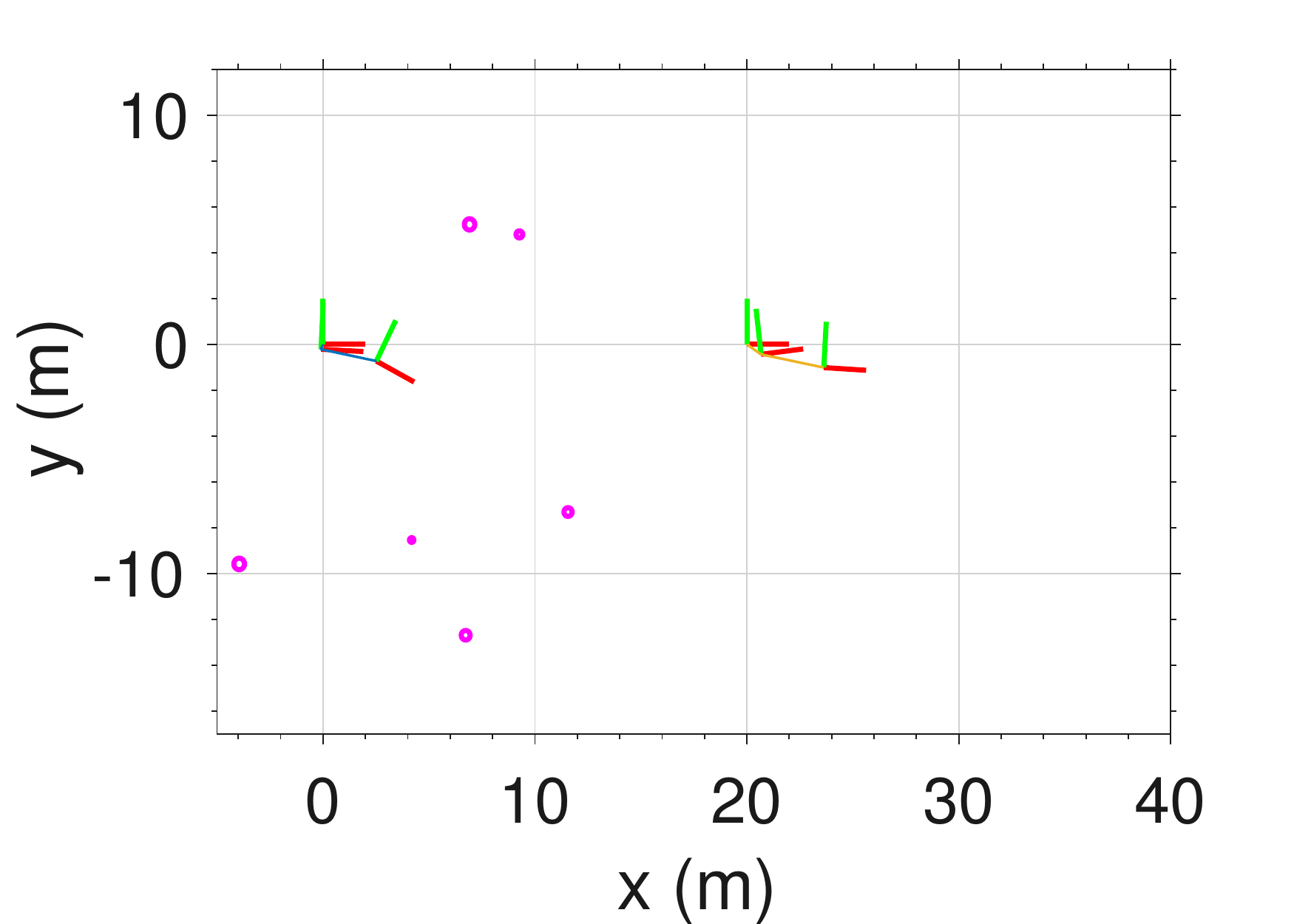}
	\caption{$t = 30$ sec\\\hspace{1in}}
	\label{fig:isam_30}
    \end{subfigure}
    \begin{subfigure}{0.42\textwidth}
	\includegraphics[width=1.0\textwidth]{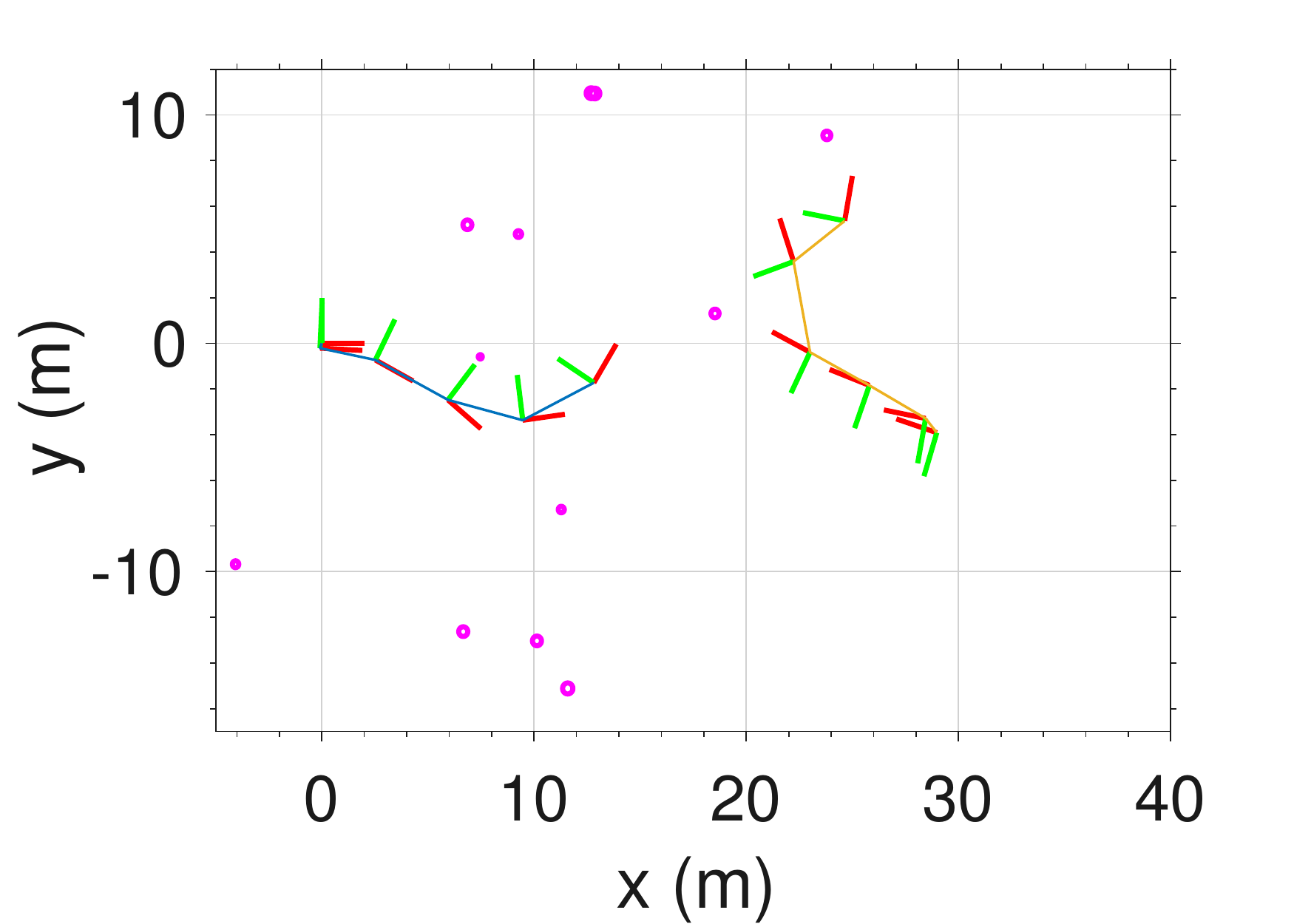}
	\caption{$t = 45$ sec\\\hspace{1in}}
	\label{fig:isam_45}
    \end{subfigure}
    \\
	\begin{subfigure}{0.42\textwidth}
	\includegraphics[width=1.0\textwidth]{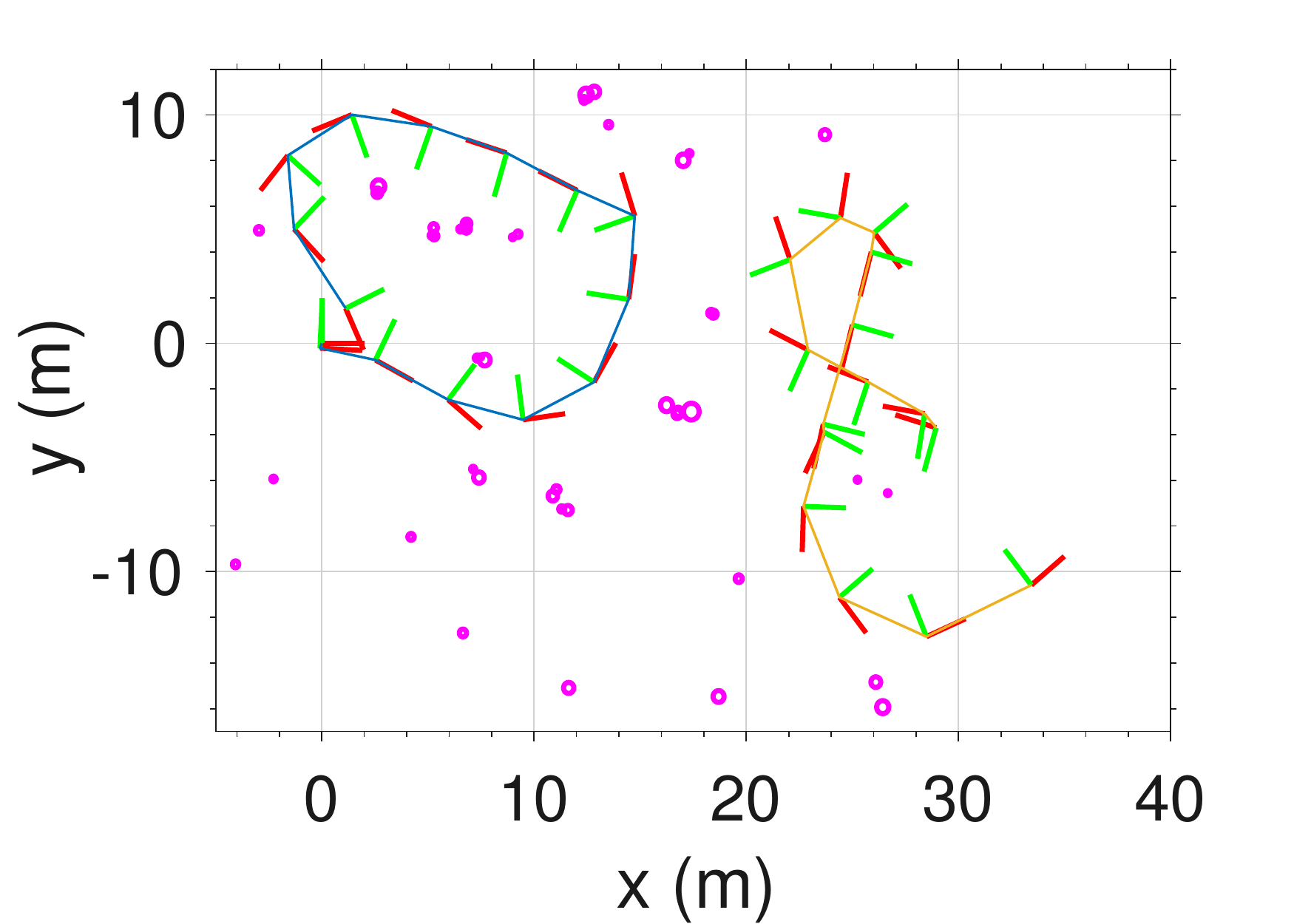}
	\caption{$t = 90$ sec\\\hspace{1in}}
	\label{fig:isam_90}
	\end{subfigure}
	\begin{subfigure}{0.42\textwidth}
	\includegraphics[width=1.0\textwidth]{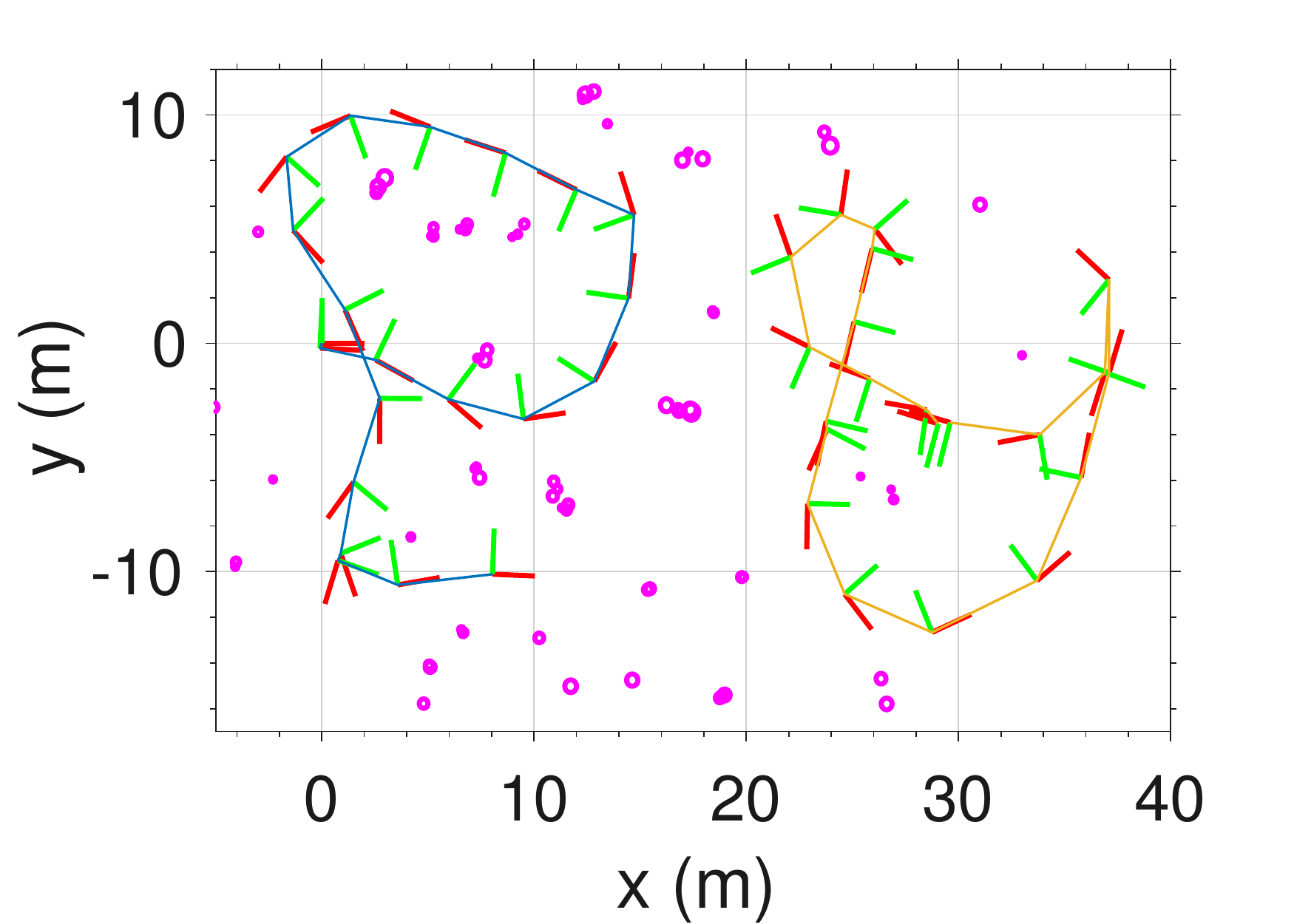}
	\caption{$t = 120$ sec\\\hspace{1in}}
	\label{fig:isam_120}
	\end{subfigure}
	\caption{Global object (tree) map on the ground station. Each coordinate frame represents the origin of a submap in the world frame. Vehicle trajectories are shown as blue and yellow lines. Each magenta circle represents a tree whose position is optimized during landmark SLAM, where the radius corresponds to the estimated tree stem radius. For cleaner visualization, we only show tree landmarks that have been observed in at least 3 submaps. At $t=30$ sec, no inter-trajectory loop closure was detected and the two vehicle trajectories are arbitrarily aligned, as shown in \textbf{(a)}. After the first inter-trajectory loop closure was established at $t = 45$ sec, the individual submaps from the two vehicles were consistently fused into the global map as shown in \textbf{(b)}-\textbf{(d)}.}
	\label{fig:isam}
\end{figure*} 

\begin{figure*}
\centering
\centerline{\includegraphics[width=0.8\textwidth]{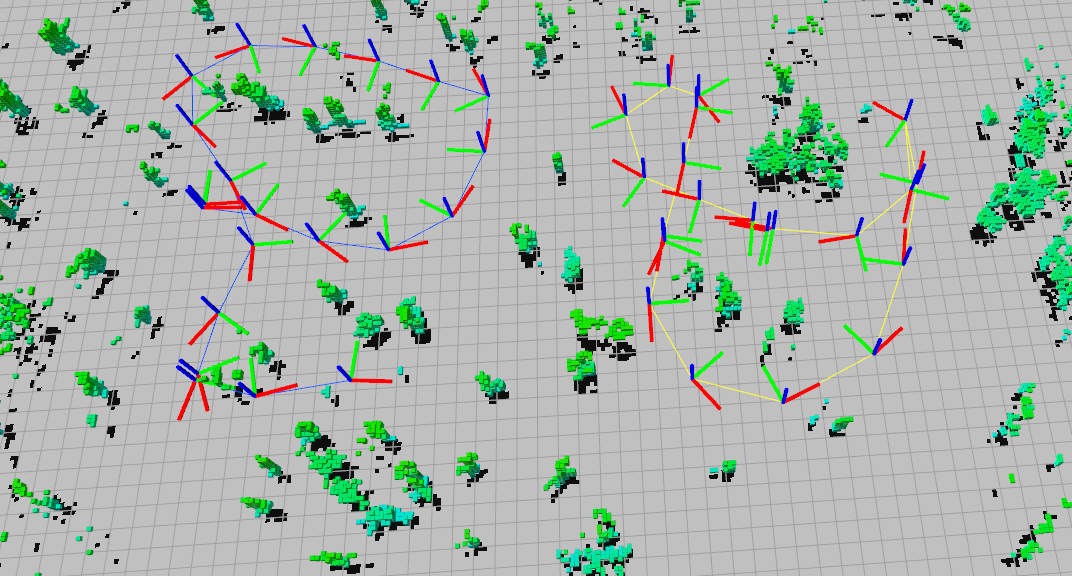}}
\vspace{0.3cm}
\caption{
Fused 3D occupancy grid and vehicle trajectories at $t=120$ sec. Each coordinate frame represents the origin of a submap in the world frame. 
Vehicle paths are shown in blue and yellow. The 3D map is represented as a voxel grid with an altitude-based colormap for occupied cells (blue to green in increasing altitude). In addition, we also show the projected 2D map. The cells are colored so that black denotes occupied cells, grey denotes free cells, and dark grey denotes unvisited cells (none in this figure). } 
\label{fig:fused_map}
\end{figure*}

In the outdoor experiments, 
our vehicles were started at different locations with unknown relative positions. 
The mission was specified
by non-overlapping search area of size $17\text{m} \times 20\text{m}$ for each vehicle.  
Each vehicle was tasked with observing the entire 
search region. Each search area was initially unknown; as the vehicles individually completed 
the coverage task, they established inter-trajectory loop closures based on 
similar configurations of trees observed during flight. The individual submaps collected during the experiment from
both vehicles were then fused real-time on the ground station.
The progress of the planners throughout the flight is shown in Figure \ref{fig:real_trajectory} and 
matching onboard camera images are shown in Figure \ref{fig:real_trajectory_gopro}. 

Since the proposed planner preferred frontiers with smaller orientation change, we observed that the resulting flight trajectory exhibited bias towards a natural spiral pattern. 
Vehicle~1 completed the exploration mission after 122 seconds, with an average flight speed of $2.66$~m/sec; vehicle~2 completed the mission after 135 seconds, with an average flight speed of $2.65$~m/sec. 
Note that due to measurement noise, the calculated average speeds were higher than the maximum allowed speed ($2.0$~m/sec) even after applying a low-pass filter to the raw velocity measurements.

Figure~\ref{fig:isam} shows the evolution of the global object (tree) map on the ground station. 
For cleaner visualization, we only show tree landmarks that have been observed in at least 3 submaps. 
At $t=30$ sec (Figure~\ref{fig:isam_30}), no inter-trajectory loop closure was detected and the two vehicle trajectories were arbitrarily aligned. 
The first inter-trajectory loop closure was established at $t = 45$ sec and map fusion happened subsequently (Figure~\ref{fig:isam_45}).
Afterwards, all newly received submaps from the two vehicles were consistently fused into the global map (Figure~\ref{fig:isam_90}-\ref{fig:isam_120}).
In addition, we also rendered the fused global 3D occupancy grid at $t=120$ sec, by aligning all submaps in the world frame using their optimized origins in Octomap \citep{hornung13auro}.
The resulting global map is shown in Figure~\ref{fig:fused_map}. Note that in practice, constructing the global occupancy grid would require vehicles to transmit their full local occupancy grids which could be communication intensive. In contrast, our tree-based map representation shown in Figure~\ref{fig:isam} captures all the essential geometric information in each submap and is still lightweight enough for low-bandwidth communication.

\Yedit{}
{
In Figure~\ref{fig:isam_120},  while we successfully fused multiple submaps into a consistent global map,   
the fused map nevertheless contains duplicate trees. 
For example, some overlapping magenta circles correspond to a single tree in the actual forest.
This issue exemplifies the trade-off between \emph{precision} and \emph{recall} 
as two performance critical metrics for data association.
In our system, we use conservative thresholds when matching trees within and across submaps.
While this approach allows us to achieve high precision, 
it also results in degrading recall (i.e., missing certain associations),
which ultimately adds duplicate trees to the map. 
However, we note that although missing associations is undesirable, it does not impact the geometric consistency of inferred loop closures. 
This is because to infer the relative transformation between two submaps in 2D, at minimum one only needs to identify two correct landmark associations. 
Therefore, although our algorithm sometimes failed to associate \emph{all} trees between two submaps, it still associated enough of them to correctly infer the underlying relative transformation, which is the most critical factor for collaborative SLAM and map fusion. 
Still, in practice 
we would like to remove any duplicate trees in order to maximize the accuracy of the final map. 
This can be done by further optimizing upstream modules (e.g., tree detection and pairwise association), or by performing an additional step to fuse overlapping landmarks after collaborative SLAM, which we leave to future work.
}

\begin{figure}
\centerline{\includegraphics[width=0.50\textwidth]{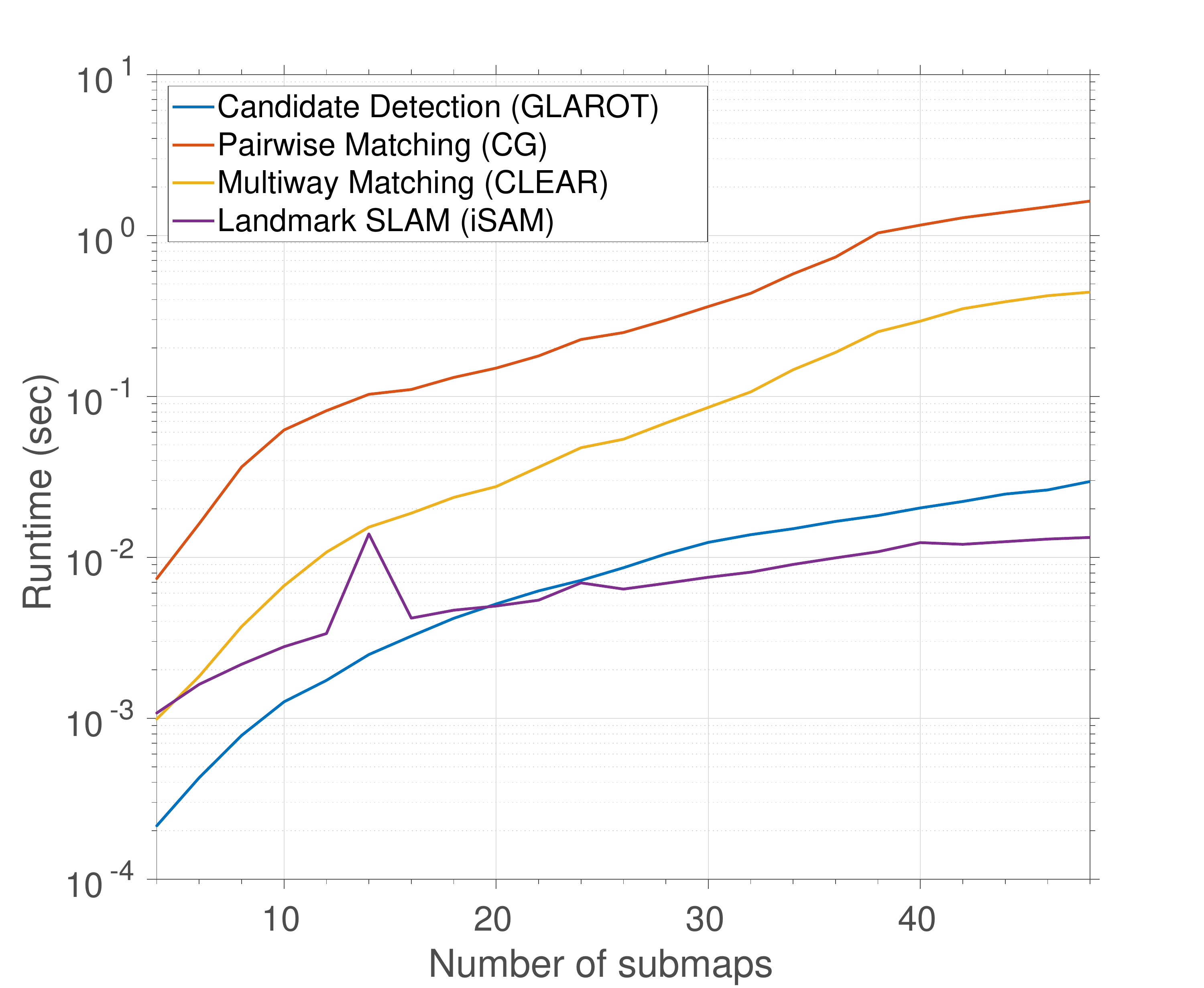}}
\caption{Total runtime of each module in the proposed CSLAM pipeline: GLAROT, CG matching, CLEAR, and iSAM, each as a function of the total number of submaps received at the ground station. 
\Yedit{
All modules demonstrated satisfactory speeds. We note that CLEAR could be further acclerated by implementing the block SVD method suggested by \cite{FathianCLEAR2019}.}
{}}
\label{fig:runtime}
\end{figure}

\begin{figure}
\centerline{\includegraphics[width=0.50\textwidth]{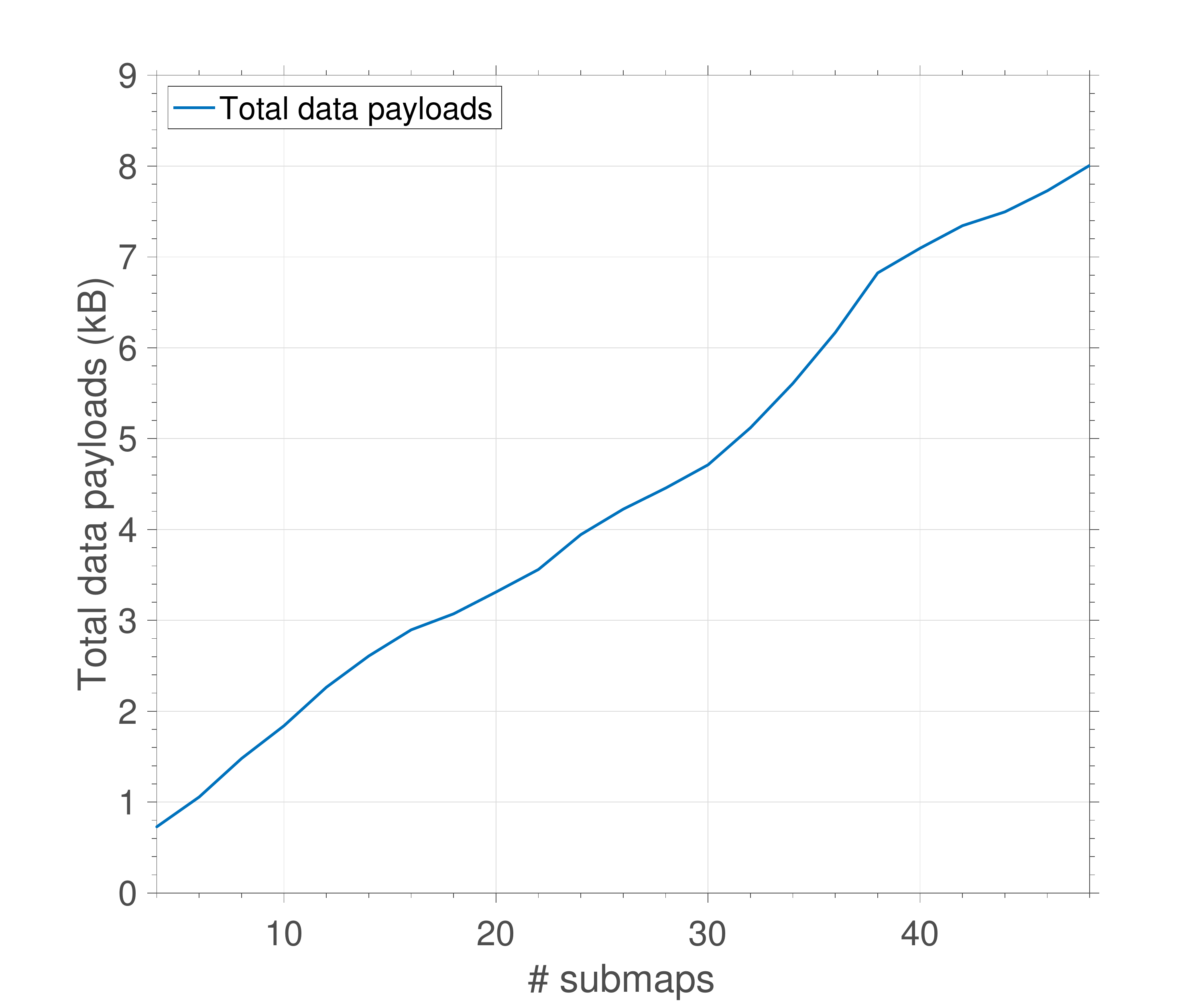}}
\caption{Data payloads, in kilobytes (kB), received at the ground station as a function of the number of submaps. Our calculation did not include additional communication incurred by network protocol overhead. In this figure, the maximum payload is $8$kB corresponding to a total of $48$ submaps. }
\label{fig:payload}
\end{figure}

To further demonstrate the efficiency of the proposed CSLAM framework, 
we also evaluated the incurred runtime and data payloads.
Figure~\ref{fig:runtime} shows the total runtime of GLAROT (Section~\ref{sec:glare}), 
CG matching (Section~\ref{sec:cg}), CLEAR (Section~\ref{sec:clear}), and iSAM (Section~\ref{sec:isam}), each as a function of the number of submaps received at the ground station. We observed that all modules demonstrated satisfactory speeds; in particular, GLAROT and iSAM were computationally very efficient and hence capable of running in real-time. In comparison, the pairwise and multiway data association steps (using CG matching and CLEAR, respectively) incurred higher computation costs.
\Yedit{
In practice, however, CG matching can be made real-time by storing and reusing previously computed matches. 
On the other hand, the CLEAR algorithm could also be accelerated by implementing a block singular value decomposition (SVD) method that exploits the separable structure of the underlying data association graph \citep{FathianCLEAR2019}.
Still, as the number of submaps grows, we note that multiway association is likely to become the computation bottleneck of the overall system. Thus, reducing the computational cost of this step, via an incremental algorithm that can reuse previously computed results, remains an important avenue for future work.}
{
In principle, CG matching has exponential complexity due to the need to solve a maximum clique problem. In our specific application, however, our use of object-level representations and aggressive culling drastically reduce the input size to CG matching (each submap typically contains less an than 30 tree landmarks). As a result, in our outdoor experiment each matching requires at most $0.015$~sec, making it possible to process incoming submaps in real-time.
On the other hand, the runtime of CLEAR eventually exceeded $0.5$~sec as more submaps are added.
We currently use CLEAR as a batch technique to periodically optimize global data association. 
An important avenue for future research is the development of incremental methods that is able to reuse previously computed results. 
}

Finally, Figure~\ref{fig:payload} shows the total data payloads received at the ground station as a function of the number of submaps. Our calculation did not include additional communication incurred by network protocol overhead. Note that the growth in data payloads was not exactly linear with respect to the number of submaps. This is because each submap contained different number of trees depending on the specific area it covered. Overall, the proposed tree-based map representation incurred minimal communication (at most $8$kB with $48$ submaps) demonstrating its usefulness for low-bandwidth and real-time communication during search and rescue.

\subsection{Data Association Results}
\label{sec:data_association_results}
\begin{figure}[t]
\centering
\begin{subfigure}{0.50\textwidth}
\includegraphics[width=1.0\textwidth,trim={0cm 0cm 0cm 0cm},clip]{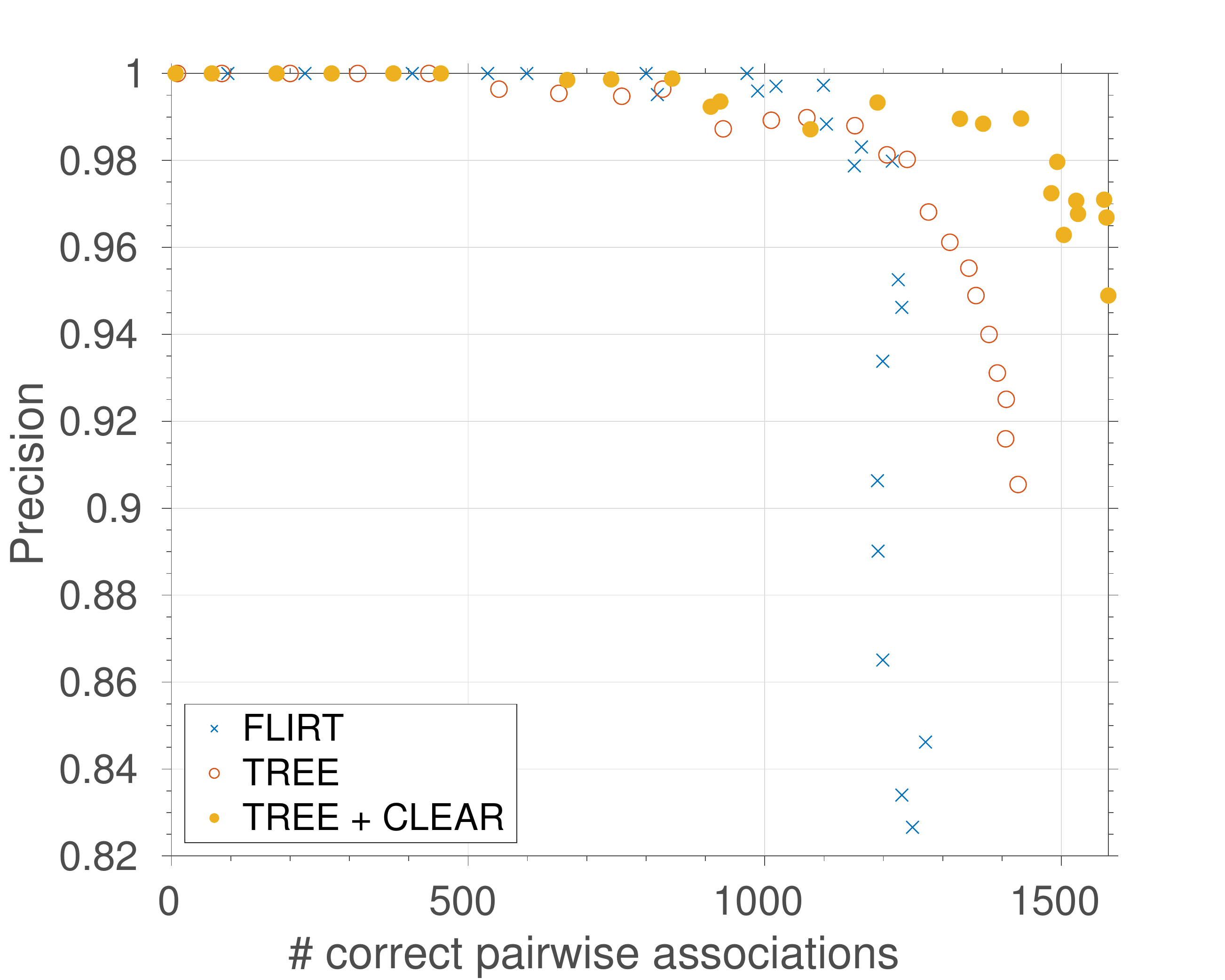}
\end{subfigure}
\caption{Precision vs. number of correct pairwise associations (unnormalized recall).
CG matching with varying tolerance threshold $\epsilon_\text{CG}$ were
performed to obtain the data points for FLIRT and trees.
Precision is computed as the percentage of correct pairwise associations within all proposed associations.
In the forest, the proposed tree detection algorithm unsurprisingly outperformed FLIRT features.
Furthermore, CLEAR was able to improve both precision and recall by simultaneously rejecting false associations and suggesting more correct associations.}
\label{fig:pvr}
\end{figure}

\begin{figure}[t]
\centering
\begin{subfigure}{0.24\textwidth}
\includegraphics[width=1\textwidth]{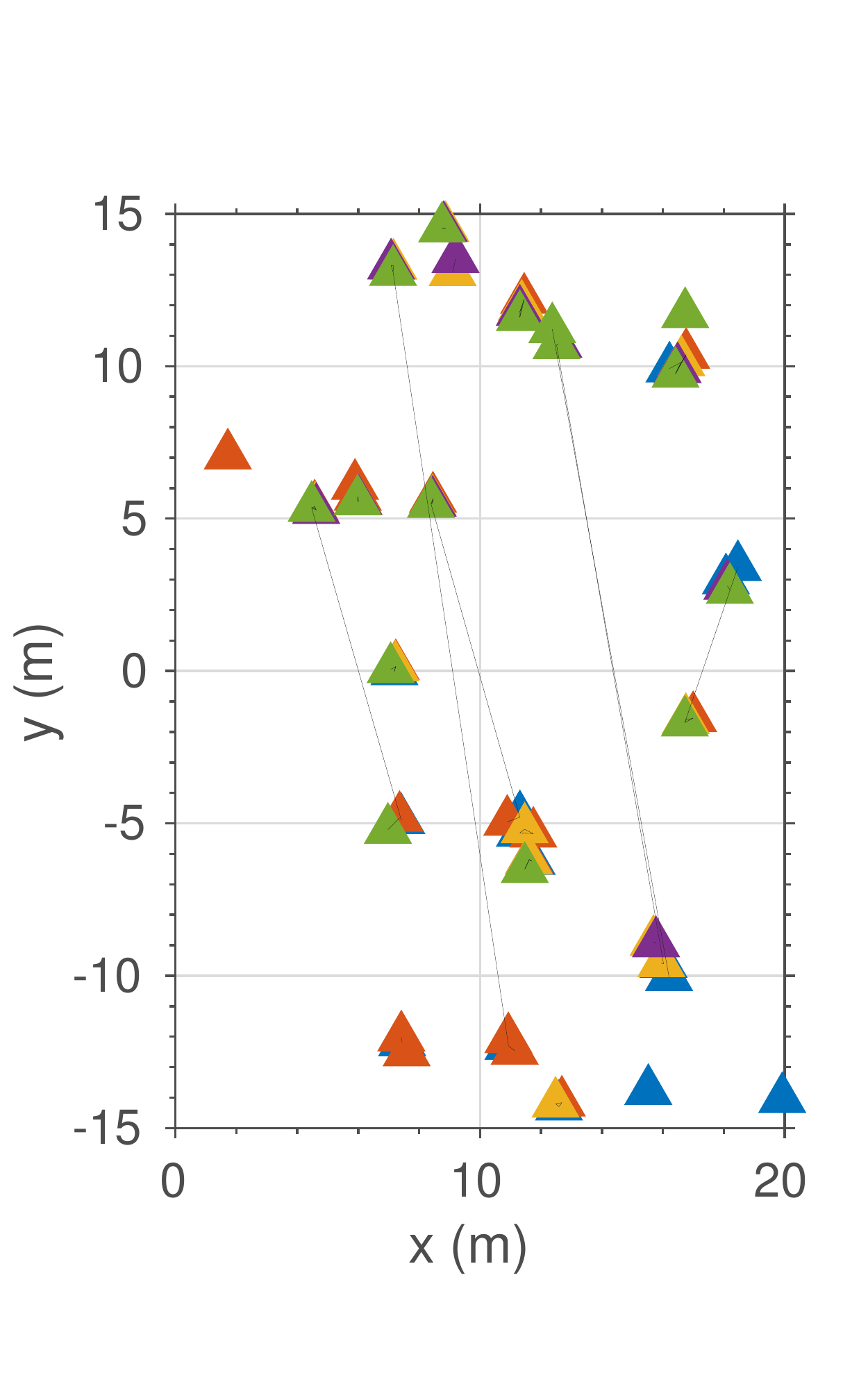}
\caption{Before CLEAR}
\label{fig:clear_input}
\end{subfigure}
\hfill
\begin{subfigure}{0.24\textwidth}
\includegraphics[width=1\textwidth]{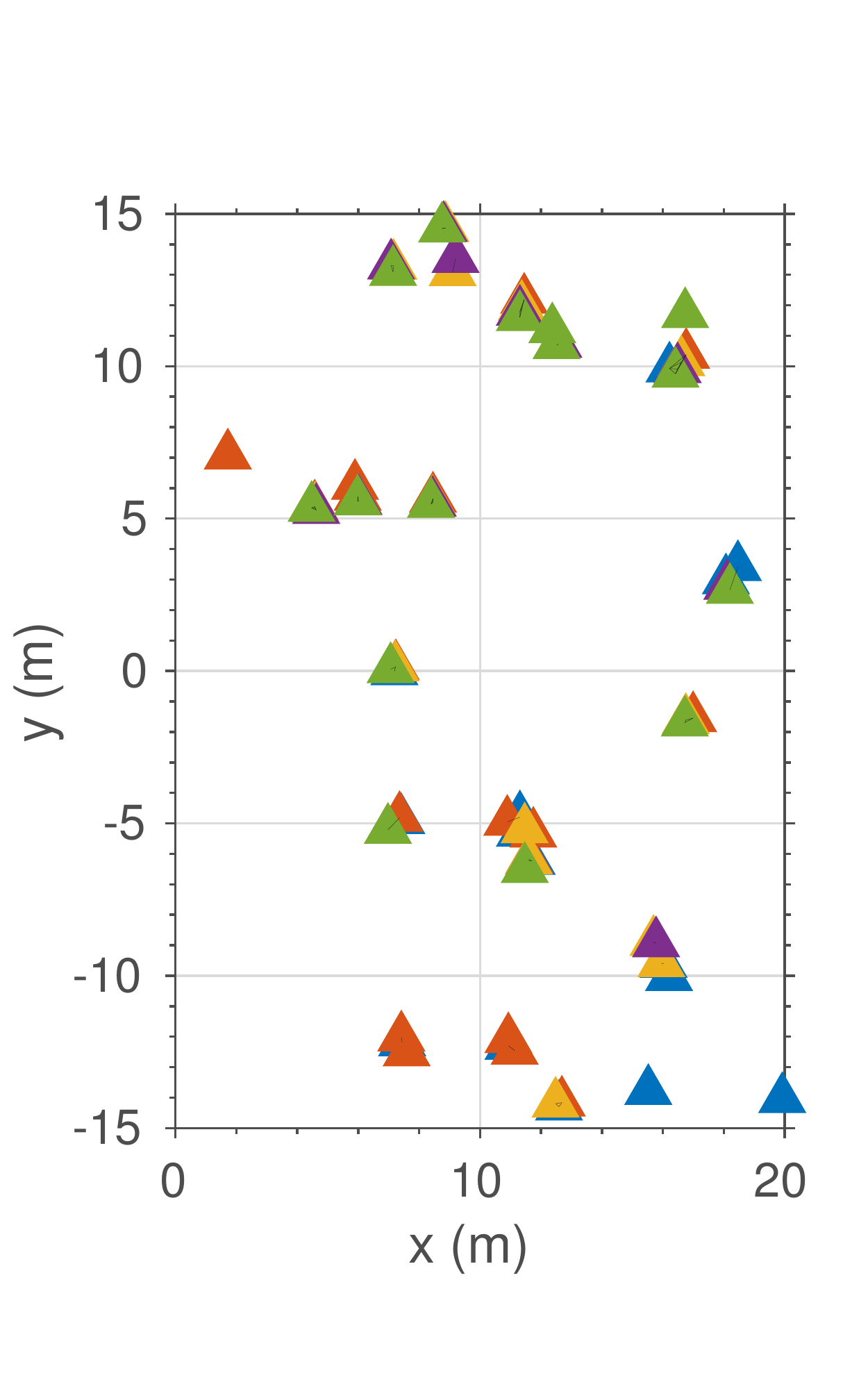}
\caption{After CLEAR}
\label{fig:clear_output}
\end{subfigure}
\caption{Data association graph before and after optimizing with CLEAR. Each triangle represents a tree colored according to the submap it belongs to. Two trees from different submaps are connected by an edge if they are matched in the given data association. By jointly optimizing over all pairwise matches, 
CLEAR was able to reject false associations that passed the initial stage of pairwise association.}
\label{fig:clear_experiment}
\end{figure}

Post-flight, we performed additional offline evaluations of the proposed pairwise and multiway data association algorithms (Section~\ref{sec:cg}-\ref{sec:clear}) using data collected by 
vehicle~2 (yellow trajectory in Figure~\ref{fig:fused_map}). For this experiment, GLAROT (Section~\ref{sec:glare}) was disabled and loop closure was attempted on every pair of submaps received at the ground station.
To benchmark our methods, we also implemented the approach in \citep{GiamouIROS2017} which uses stabilized FLIRT features obtained from DP-means clustering. 
Ground truth (e.g., GPS) data was not available in our forest datasets;
instead, we used estimates provided by onboard EKF to verify proposed pairwise associations.
Specifically, a pair of matched trees was declared to be a true match
if their positions as estimated by EKF in the world frame were sufficiently close. 
For short-duration flights, we expect this metric to be relatively accurate.
\Yedit{}{For longer flights, the accuracy of this metric will degrade as drifts accumulate in the EKF estimate.
A better evaluation method should utilize more accurate source of information (e.g., GPS)
or uncertainty-aware metrics (e.g., mahalanobis distance), which we leave to future work.}

Figure~\ref{fig:pvr} shows the resulting precision and 
\Yedit{number of correct pairwise associations (unnormalized recall)}{unnormalized recall (number of correct pairwise associations)}\footnote{
We do not show recall as the total number of correct pairwise associations in this dataset is unknown.},
\Yedit{}{which are two commonly used performance metrics when evaluating the quality of data association. }
For each submap, 
an equal number of stabilized FLIRT features \citep{GiamouIROS2017}
and tree landmarks were extracted. 
CG matching (Section~\ref{sec:cg}) with varying tolerance threshold $\epsilon_\text{CG}$ were
performed to obtain the resulting data points for FLIRT and trees.
For trees, we further optimized the initial pairwise associations with the CLEAR algorithm (Section~\ref{sec:clear}).
In the forest, the proposed tree detection algorithm unsurprisingly outperformed stabilized FLIRT features.
Furthermore, CLEAR was able to improve both precision and recall by simultaneously rejecting false associations and suggesting more correct associations. This result demonstrates the usefulness of cycle consistent multiway association as an additional step for CSLAM.

To provide more insights on the benefits of cycle consistent multiway matching, Figure~\ref{fig:clear_experiment} shows
part of the data association graph before and after optimizing with CLEAR.
Each triangle represents a tree colored according to the submap it belongs to.
Two trees from different submaps are connected by an edge
if they are matched in the given data association.
For visualization purpose, all submaps are aligned in the world frame, i.e.,
each object is drawn at the corresponding position estimated by EKF.
Thus, objects that correspond to the same tree in the forest should appear close to each other.
Due to perceptual aliasing,
the initial associations from CG matching contained several wrong correspondences, shown as the long edges in Figure~\ref{fig:clear_input}.
By jointly optimizing over all pairwise matches, 
CLEAR was able to reject these outliers as shown in Figure~\ref{fig:clear_output}.
Furthermore, the resulting data association is guaranteed to be cycle consistent. 
This means that the graph shown in Figure~\ref{fig:clear_output} only contains disjoint cliques
which makes subsequent map fusion possible;
see Section~\ref{sec:clear}.

\section{Conclusions and Future Work}
\label{sec:insights}

We presented a collaborative exploration and mapping system for multi-UAV search and rescue under the forest canopy. Our UAVs were equipped with onboard autonomy that reliably performed sensing, pose estimation, local mapping, and exploration planning. CSLAM was performed at a central ground station. To cope with the limited communication bandwidth, we used a map compression scheme that compressed onboard dense occupancy grids into sparse tree-based maps before transmission.
During CSLAM, we used the relative positions and orientations between multiple trees as a unique signature to detect loop closures. To further improve both precision and recall during data association, we proposed cycle consistent multiway matching as an additional stage in our CSLAM pipeline. Extensive experiments were carried out to validate the proposed system, including a real-world flight demonstration at a forest in NASA LaRC.

\Yedit{}{
While we have demonstrated the use of multiple UAVs for search and rescue, many aspects of our system can be further improved.
In this work, we have not explicitly addressed the issue of outliers during data association and loop closure detection. 
In our outdoor experiment, we rarely observed outlier loop closures due to the conservative thresholds used in CG matching.
Nonetheless, under higher noise regimes (e.g., in denser forests with more branches and leaves), errors during data association and loop closure are more likely to happen. 
In these scenarios, one needs to implement outlier rejection schemes for both inner-trajectory and inter-trajectory loop closures. 
As an example, the robust map fusion procedure in \citep{JoshuaICRA2018} can be directly applied on top of our current system.
} 

\Yedit{While }{On the planning side, currently each vehicle performs frontier-based exploration within its own search region.}
While the dynamics-aware frontier selection was able to produce smoother and more time efficient search behavior, we also observed that its inherent bias to follow the current velocity vector sometimes results in the vehicle flying past a small patch of unobserved space, and having to return later in the mission. 
Possible ways of addressing this problem include encouraging the planner to prioritize small patches of unobserved space, or incorporating longer-horizon planning during exploration. 
\Yedit{An example is shown in Figure~\ref{fig:real_trajectory}, where the second vehicle missed the tear-dropped
shaped section of unknown space.}
{In addition, currently each of our vehicles performs independent exploration within its assigned region, without much interactions with others
(except during collaborative SLAM). 
Since the search regions do not overlap with each other, 
explicit coordination is not required to complete the overall exploration task.
Nevertheless, our system could still benefit from higher-level coordination among the robots to be more flexible at, e.g., 
assigning vehicles to different search areas at different times. 
}

\Yedit{}{Finally, most procedures in our system currently take place in the projected 2D map. 
In our experiments, we observed that the 2D assumption is generally sufficient for sparse forests. 
Nevertheless, such assumption could become restrictive in denser environments, for example when feasible 2D paths no longer exist. 
We note that most modules in our system can be extended to 3D at the expense of increased computation costs, without requiring many modifications.
Such an extension will also be considered for future work.}

\subsection*{Acknowledgements}
The authors gratefully acknowledge Matthew Giamou for early work on this project, and Chester Dolph, Brian Duvall, and Zak Johns for assistance during the outdoor experiments.
The authors also would like to thank Kasra Khosoussi and Kaveh Fathian for insightful discussions during the preparation of this manuscript,
\Yedit{}{as well as two anonymous reviewers for their constructive feedback that further improved this work}.

This work was supported in part by the NASA Convergent
Aeronautics Solutions project Design Environment for Novel
Vertical Lift Vehicles (DELIVER), by ONR under BRC award N000141712072, and by
ARL DCIST under Cooperative Agreement Number W911NF-17-2-0181.

\bibliographystyle{SageH}
\bibliography{ijrr19}
\end{document}